\documentclass{article}

\PassOptionsToPackage{numbers, compress}{natbib}
\usepackage[preprint, nonatbib]{neurips_2025}
\makeatletter
\renewcommand{\@noticestring}{}
\makeatother

\usepackage[utf8]{inputenc}
\usepackage[T1]{fontenc}
\usepackage{microtype}
\usepackage{amsmath, amssymb, amsfonts}
\usepackage{booktabs, array, multirow, longtable, adjustbox, tabularx, makecell}
\usepackage[table]{xcolor}
\usepackage{hyperref}
\usepackage{graphicx, float, caption}
\usepackage{enumitem}
\setlist[itemize]{topsep=2pt, itemsep=1pt}

\title{Turn-Averaged SAEs for Feature Discovery and Long-Context Attribution}

\author{
  Kevin Der \\
  Anthropic Fellows Program
  \And
  Harish Kamath \\
  Anthropic
  \And
  Ben Thompson \\
  Anthropic
}

\graphicspath{{figures/}}

\begin{document}
\maketitle

\begin{abstract}
Sparse autoencoders (SAEs) have become a useful tool for extracting interpretable features in language models. However, standard SAE architectures operate on individual token activations, meaning that the number of active features scales linearly with context length, and studying long model transcripts becomes difficult. We introduce turn-averaged SAEs, which represent a single Human or Assistant turn with a fixed number of features by learning to reconstruct the average model activation across the turn. We find that turn-averaged features describe a single turn's high-level characteristics more completely than per-token features when judged by an LLM. We also demonstrate that turn-averaged SAEs greatly simplify common downstream uses of SAEs like attribution graphs. Broadly, turn-averaged SAEs make interpretability techniques practical at long context lengths.
\end{abstract}

\section{Introduction}
\label{sec:intro}

Sparse autoencoders decompose model activations into sparse, interpretable features, and have found broad application in circuit analysis, behavioral steering, and safety monitoring. Existing SAE methods typically operate on individual token activations, producing feature data that scales linearly with sequence length. Since many interpretability techniques are ultimately tools for analysis, per-token scaling presents a practical limitation. For example, consider the goal of understanding how SAE features affect a model's output. The features that fire on a given span of text might appear to have a compelling relevance to the output, but their activation values alone do not establish causality. Attribution graphs address this by estimating first-order effects between features across layers, but at per-token resolution, long contexts and multi-turn conversations produce graphs that are cumbersome to inspect and comprehend.

Beyond the scaling challenge, interpretability researchers often wish to understand why a model produced a particular response. As such, it is valuable to have tools which identify the high-level properties that characterize a response as a whole. Examples of such properties include topic, style, persona, and safety-relevant traits such as adversarial prompt detection. At least some such signals, including persona-related ones, are spread across many tokens rather than localized to a few positions (Chen et al., 2025). Many per-token SAE features, in contrast, act as detectors for specific words or syntactic patterns, firing at only a small number of token positions. Because per-token SAEs reconstruct each position's activation independently, nothing in the training objective compels such high-level characteristics to separate from token-level detail in the learned features.

We propose a simple modification: training SAEs on the mean hidden state across a token span rather than on individual token activations. This addresses the scaling challenge by reducing the unit of analysis from individual tokens to spans. The technique is straightforward to implement, as averaging preserves the dimensionality of the input and so existing SAE architectures and training code require no modification. Although our experiments focus on human-assistant conversation turns, the approach generalizes to spans of arbitrary length, as our long-context results show (Section~\ref{sec:feature_discovery}).

Our contributions are the following:

\begin{enumerate}
  \item We show that training SAEs on turn-averaged activations produces features that capture high-level characteristics of a turn, and compare these against per-token features using both qualitative and quantitative evaluations.
  \item We introduce a nested SAE architecture that jointly trains turn-averaged and per-token features within a single model using a Matryoshka-like loss.
  \item We adapt attribution graphs to the turn level, enabling compact analysis of multi-turn conversations. Edge weights are validated through intervention experiments, as well as completeness and replacement sufficiency metrics that measure how thoroughly the features explain the model's computation.
\end{enumerate}

\section{Background \& Related Work}
\label{sec:background}

\subsection{Sparse autoencoders for interpretability}

Bricken et al.\ (2023) and Cunningham et al.\ (2023) demonstrated that sparse autoencoders extract interpretable features from language model activations. These features can be used to steer model behavior, including along safety-relevant dimensions, and extend to production-scale models (Templeton et al., 2024). Various sparsity mechanisms have since been developed, including BatchTopK (Bussmann et al., 2024), which we use in this work. Feature descriptions are typically generated by prompting an LLM with a feature's top-activating examples (Bills et al., 2023; Gao et al., 2024).

\subsection{Temporal and multi-scale representations}

\textbf{Temporal SAEs.} Bhalla et al.\ (2025) introduce Temporal SAEs, which apply a contrastive loss to a designated subset of features, encouraging those features to have similar activations on adjacent tokens within the same sequence. The contrastive loss favors features whose activations vary smoothly across the token dimension, which the authors associate with semantic rather than syntactic content. We target a different property: rather than encouraging smooth activation across adjacent tokens, we train on the mean hidden state across an entire span, producing features that characterize the span as a whole regardless of how individual token activations vary within it. Our approach is motivated by the observation that conceptual features from a per-token SAE do not necessarily fire strongly on adjacent tokens --- it is common for such features to activate at scattered positions across a context rather than varying smoothly. Lubana et al.\ (2025) observe that standard SAEs assume an i.i.d.\ prior over token positions, ignoring temporal structure in model activations, and propose separating activations into a predictable component derived from past context and a novel component that captures what the current token adds.

\textbf{Matryoshka SAEs.} Matryoshka Representation Learning (Kusupati et al., 2022) trains nested subspaces of increasing dimension, each independently functioning as a representation. Bussmann et al.\ (2025) apply this principle to SAEs, using nested dictionaries where smaller partitions capture general concepts and larger ones capture specifics. Temporal SAEs use a Matryoshka-style partition to separate the features subject to the contrastive loss from the remaining per-token features. A similar formulation can be adapted to compel features within a single SAE to represent either high-level characteristics or per-token detail, as our work explores.

\subsection{Information concentration across token positions}

Turn-averaging treats every token position equally when computing the mean hidden state; however, some work indicates that language models do not always distribute information evenly across token positions. Gurnee \& Tegmark (2023) find that the hidden state at a period token encodes summary information about the preceding sentence, beyond what is present at other positions. Similarly, Geva et al.\ (2023) find that for multi-token entity names, models concentrate information about the entity at the name's last token. Further investigation into how information concentrates at specific tokens and how this interacts with turn-averaging is a relevant direction that we leave to future work.

\subsection{Attribution and circuit analysis}

Ameisen et al.\ (2025) introduce attribution graphs, which for a given input quantify how features at each layer affect features at subsequent layers and the model's output. Each node represents an active feature or an output logit, and directed edges estimate the effect of one node's activation on another using a first-order approximation. Edge weights are computed using a linearized backward pass through the model, capturing how much each source node's activation contributes to a target node. Completeness and replacement sufficiency metrics measure how thoroughly the feature nodes explain the model's computation vis-\`a-vis unexplained error nodes. Such graphs are typically built using SAE features, so their size depends on the number of active features and the granularity of the activations they describe. When the SAE operates at per-token resolution, the resulting graphs have node counts that scale with sequence length. We adapt attribution graphs to turn-averaged features, producing nodes per turn per layer rather than per token per layer, yielding compact graphs whose nodes represent model behavior in terms of high-level characteristics. As an alternative approach, Arora et al.\ (2026) demonstrate that circuit tracing can also be performed directly in the neuron basis without learned features.

\section{Method}
\label{sec:method}

\subsection{Turn-averaged SAEs}

We train sparse autoencoders on the mean hidden-state activation across a span of tokens, rather than on individual per-token activations. Such SAEs, which we refer to as turn-averaged SAEs, use a training objective that is standard SAE reconstruction loss --- the only difference from a per-token SAE is the input.

Given a span T of tokens, let $x_t$ $\in$ $\mathbb{R}^{d_{\text{model}}}$ be the residual stream at position t. Compute the turn average:

\begin{equation*}
  \bar{x} = (1/|T|) \sum _{t\in T} x_t
\end{equation*}

The SAE is trained on {$\bar{x}$} with standard reconstruction loss:

\begin{equation*}
  \mathcal{L} = \|\bar{x} - \text{dec}(\text{enc}(\bar{x}))\|^2
\end{equation*}

where $\text{enc}(\bar{x}) = \text{BatchTopK}(W_{\text{enc}} (\bar{x} - b_{\text{dec}}) + b_{\text{enc}})$ and $\text{dec}(z) = W_{\text{dec}} z + b_{\text{dec}}$.

Because the mean of a set of vectors in $\mathbb{R}^{d_{\text{model}}}$ has the same shape as any individual vector, the SAE architecture and training code require no changes to accept turn-averaged samples. In our experiments, each span corresponds to a single turn in a multi-turn chat transcript. The technique applies to any segmentation of a token sequence, including spans of arbitrary length.

\textbf{Setup.} We train on Qwen-2.5-7B-Instruct \cite{qwen2024} ($d_{\text{model}}$ = 3584) using LMSYS-Chat-1M \cite{zheng2023} as the training corpus. All SAEs use BatchTopK with $d_{\text{sae}}$ = 32,768 and $k = 128$. Training takes approximately 1.5 hours on 4$\times$H100 for one epoch (\textasciitilde{}1.58M turn-averaged samples), compared to 266M tokens for a per-token SAE on the same data. Additional training details are in Appendix~\ref{app:training}.

\textbf{Intuition.} Why does averaging compel the SAE to learn high-level features? Many per-token SAE features are low-level --- e.g., single-token detectors and lexical or syntactic patterns --- because much of a token's representation must encode where it is in the context and what comes next. When averaging activations over many positions, these per-token characteristics are smoothed away because they vary across tokens. If we consider the hidden
states across a span as a signal over the token dimension, we expect that broad concepts like
topic and behavioral properties are often reflected across many positions and correspond to low-frequency
components of this signal. Taking the mean acts as a low-pass filter that captures these components
while also attenuating per-token noise. The high-frequency component that is discarded corresponds
to detail specific to individual tokens.

\subsection{Hybrid architectures}
\label{sec:hybrid}

Turn-averaged SAEs capture high-level features but cannot reconstruct per-token detail --- the averaging discards the high-frequency component of the signal. A natural question is whether a single model can learn features that separately capture both components: the low-frequency turn-level structure and the fine-grained per-token residual. We explore two architectures: a decoupled model that trains separate SAEs on orthogonal signal components, and a nested model that trains a single SAE with a Matryoshka-like loss.

\subsubsection{Decoupled SAE}

Per-token activations can be written as the sum of the mean hidden state across the turn and a residual:

\begin{equation*}
  x_t = \bar{x} + (x_t - \bar{x})
\end{equation*}

The decoupled model trains two independent SAEs: a turn-averaged SAE on $\bar{x}$ and a per-token SAE on the residual ($x_t$ $-$ $\bar{x}$). Each uses standard reconstruction loss on its respective input:

\begin{gather*}
  \mathcal{L}_{\text{TA}} = \|\text{dec}_c(\text{enc}_c(\bar{x})) - \bar{x}\|^2 \\
  \mathcal{L}_{\text{PT}} = \|\text{dec}_f(\text{enc}_f(x_t - \bar{x})) - (x_t - \bar{x})\|^2
\end{gather*}

Combined per-token reconstruction sums the two decodings:

\begin{equation*}
  \hat{x}_t = \text{dec}_c(\text{enc}_c(\bar{x})) + \text{dec}_f(\text{enc}_f(x_t - \bar{x}))
\end{equation*}

Note that the turn-averaged features contribute the same reconstruction to every token in the turn, while the per-token features vary by position. The two SAEs are fully decoupled --- they share no gradients and no parameters. The turn-averaged features are trained entirely on the mean component of the signal, and the per-token features specialize on the residual. Each SAE uses BatchTopK with $k = 64$, yielding 128 total features per token, matching the feature budget of the other architectures.

However, full decoupling has significant drawbacks. The separation of the two underlying models does not allow typical analysis techniques such as attribution --- the two feature sets are causally disconnected because the per-token SAE's input $h_t - \bar{x}$ is invariant to the uniform shifts produced by ablating turn-averaged features. Also, to compare against other architectures at a fixed top-k budget, that capacity must be split across the two models, adding an additional hyperparameter and yielding worse reconstruction than a single model at the same total k.

\subsubsection{Nested SAE}

Noting the shortcomings of a decoupled model, we are motivated by the following high-level goal: in a single SAE, can a subset of features optimally reconstruct the mean of the activations while the full set of features reconstructs the per-token activations? That is, some features are compelled to capture the low-frequency component of the signal, while the full set of features captures the entire signal. To this end, we divide the SAE feature space into an inner partition [0:h] and an outer partition [h:$d_{\text{sae}}$], with the following loss:

\begin{equation*}
  \mathcal{L} = \sum _t \|\text{dec}(\text{enc}(x_t)) - x_t\|^2 + \alpha  \cdot  \|\text{dec}_h(\text{enc}(\bar{x})[0:h]) - \bar{x}\|^2
\end{equation*}

where $\text{dec}_h$ uses only the first h decoder directions, and $\text{enc}(\bar{x})[0{:}h]$ masks features $[h{:}]$ to zero before TopK. Unlike the decoupled model, the two partitions share a single encoder and decoder, with the loss acting as a soft constraint to distinguish turn-averaged features from per-token ones. Our loss formulation is similar to Matryoshka SAE methods \cite{bussmann2025}, with the modification that the training signal for the inner partition is reshaped using turn averaging.

The weight $\alpha$ controls the relative importance of turn-averaged reconstruction; we sweep $\alpha$ $\in$ \{0.25, 0.5, 1.0, 2.0, 4.0\} and use $\alpha$ = 1.0, which balances the two loss terms while avoiding the reconstruction degradation seen at either end of that range. In practice, our experiments use $d_{\text{sae}}$ = 32,768 and $h = 16{,}384$ --- that is, we allocate equal capacity to the two classes of features.

\subsection{Reconstruction quality}

We use normalized mean squared error (nMSE) to quantify reconstruction quality across different models:

\begin{equation*}
  \text{nMSE} = \frac{\mathbb{E}[\|x - \hat{x}\|^2]}{\mathbb{E}[\|x\|^2]}
\end{equation*}

Table~\ref{tab:reconstruction} reports nMSE on the full holdout set. The nested inner partition achieves 0.116 despite contributing to both loss terms. In terms of per-token activations, both the nested and decoupled architectures closely match the pure per-token SAE, while additionally capturing turn-level structure. We also measure turn-averaged nMSE on the per-token SAE by encoding the mean hidden state through it --- the high error (0.355) indicates that the mean is significantly off-distribution for a model trained on individual token activations.

\begin{table}[!htbp]
\centering
\small
\begin{tabular}{@{}lllll@{}}
\toprule
Model & $d_{\text{sae}}$ & k & turn-averaged nMSE & per-token nMSE \\
\midrule
Turn-averaged & 32,768 & 128 & \textbf{0.097} & 1.799 \\
Per-token & 32,768 & 128 & 0.355 & \textbf{0.162} \\
Nested & 16,384 / 16,384 & 128 & 0.116 & 0.167 \\
Nested & 8,192 / 24,576 & 128 & 0.120 & 0.169 \\
Decoupled & 16,384 / 16,384 & 64+64 & 0.128 & 0.212 \\
\bottomrule
\end{tabular}
\caption{Reconstruction quality (nMSE) on the holdout set. For nested and decoupled models, $d_{\text{sae}}$ shows the inner (turn-averaged) / outer (per-token) partition sizes. The decoupled model uses $k = 64$ per component, yielding 128 total features to match the other architectures.}
\label{tab:reconstruction}
\end{table}

\section{Qualitative Feature Exploration}
\label{sec:feature_discovery}

We compare which features activate on a given prompt across three SAE architectures: turn-averaged, per-token, and nested. The turn-averaged SAE's top features describe the high-level characteristics of a response more completely than those of the per-token architecture. The nested architecture recovers such corresponding features often, but not completely, as a portion of its capacity is allocated to per-token detail.

\textbf{Setup.} We encode holdout turns not seen during training through each SAE and compare the resulting active features. Per-token features are aggregated by summing each feature's activation across all tokens in the turn and taking the top $k = 128$, matching the value used in SAE training. The subsequent quantitative evaluation also tests max pooling, but we found by inspection that it produces feature lists that capture high-level characteristics far less reliably than summing. Automated interpretability assigns a description to each feature by prompting Claude \cite{anthropic2024} with its top 25 activating examples. For turn-averaged SAEs, the prompt consists of the turn's text with the feature's activation value. For per-token SAEs, the prompt annotates individual highly-activating tokens with their activation values. For the nested SAE, we follow the same procedure as the per-token SAE: we encode per-token activations (not the mean), making no distinction between inner and outer partition features.

\textbf{LMSYS examples.} We selected the following from 316,000 held-out assistant turns (53.2M tokens) by manually browsing the results and choosing cases where the contrast between turn-averaged and per-token features was illustrative.

\textbf{A.} \textit{``The highest number below 100 that does not contain the digit 9 is 95.''}
\begin{center}
\scriptsize
\renewcommand{\arraystretch}{1.4}
\begin{tabular}{@{}>{\raggedright\arraybackslash}p{1.5cm}>{\raggedright\arraybackslash}p{3.7cm}>{\raggedright\arraybackslash}p{3.7cm}>{\raggedright\arraybackslash}p{3.7cm}@{}}
\toprule
Category & Turn-averaged & Per-token & Nested \\
\midrule
Error recognition & "Incorrect answers to digit-exclusion number puzzles" \newline "Confidently stating incorrect mathematical and numerical facts" & (none found) & "Incorrect or hallucinated answers to word/number puzzles and constraints" \newline "Arithmetic and number system calculations with frequent errors" \\
\specialrule{0.2pt}{2pt}{2pt}
Numerical reasoning & "Numerical calculations and large number outputs" \newline "Responses discussing, identifying, or explaining numbers and digits" & "Digit sum calculations and numerology number sequences" \newline "Numerical reasoning: digits, place values, arithmetic operations, number systems" & "Explaining significance or properties of specific numbers" \\
\bottomrule
\end{tabular}
\end{center}

Turn-averaged features identify that the answer is \textit{incorrect} --- a behavioral property of the response. Per-token features find digits and numerical reasoning but miss the incorrectness entirely. The nested architecture also recovers the error-recognition concept, confirming that turn averaging captures this behavioral property.

\textbf{B.} \textit{``The neuron's electrochemical dance underneath the daily onslaught of neurotoxins -- like a firework show in your brain!''}
\begin{center}
\scriptsize
\renewcommand{\arraystretch}{1.4}
\begin{tabular}{@{}>{\raggedright\arraybackslash}p{1.5cm}>{\raggedright\arraybackslash}p{3.7cm}>{\raggedright\arraybackslash}p{3.7cm}>{\raggedright\arraybackslash}p{3.7cm}@{}}
\toprule
Category  &  Turn-averaged  &  Per-token  &  Nested \\
\midrule
Enthusiastic technical style  &  "Enthusiastic, engaging explanations of technical and factual topics"  &  "Casual, energetic first-person character speech in quotes"  &  "Enthusiastic, promotional, or upbeat tone in assistant responses" \\
\specialrule{0.2pt}{2pt}{2pt}
Creative writing  &  "Elaborate, flowery creative writing with dramatic literary prose"  &  "Poetic, lyrical, or literary creative writing passages"  &  "Poetic, emotionally expressive prose with romantic or melancholic themes" \\
\specialrule{0.2pt}{2pt}{2pt}
Figurative language  &  "Simile-based comparisons and metaphorical expressions in varied contexts"  &  "Metaphor, analogy, and simile contexts across languages"  &  "Simile and metaphor usage: 'like' or 'as if' comparisons" \\
\specialrule{0.2pt}{2pt}{2pt}
Neuroscience  &  "Factual neuroscience explanations about neurotransmitters and brain regions"  &  "Neuroscience and brain anatomy vocabulary across entire responses"  &  "Brain and neuroscience topic responses" \\
\bottomrule
\end{tabular}
\end{center}

Turn-averaged features identify four aspects of the prompt: enthusiastic technical style, creative writing, figurative language, and neuroscience. Of these, the per-token SAE misses the aspect of technical explanation when it identifies the style. The nested architecture identifies the style.

\textbf{C.} Automotive functional safety blog outline (351 tokens).
\begin{center}
\scriptsize
\renewcommand{\arraystretch}{1.4}
\begin{tabular}{@{}>{\raggedright\arraybackslash}p{1.5cm}>{\raggedright\arraybackslash}p{3.7cm}>{\raggedright\arraybackslash}p{3.7cm}>{\raggedright\arraybackslash}p{3.7cm}@{}}
\toprule
Category  &  Turn-averaged  &  Per-token  &  Nested \\
\midrule
Outline structure  &  "Structured outlines with Roman numeral hierarchical sections"  &  "Structured outline tokens: bullets, section markers, list items"  &  "Structured outlines with Roman numeral hierarchical sections" \newline "Technical engineering and mechanical systems explanations with numbered lists" \\
\specialrule{0.2pt}{2pt}{2pt}
Automotive domain  &  "Helpful automotive advice, diagnostics, and repair information"  &  (none found)  & (none found) \\
\specialrule{0.2pt}{2pt}{2pt}
Automotive safety standard  &  (none found)  & (none found) &  "Safety, security, and risk management explanations" \\
\specialrule{0.2pt}{2pt}{2pt}
Regulatory compliance  &  "Regulatory compliance, safety standards, and industry guidelines content" \newline "Explaining technical industry standards and regulatory frameworks"  &  (none found)  & (none found) \\
\specialrule{0.2pt}{2pt}{2pt}
Generic / lexical  &  (none found)  &  "Production process and technical subject matter descriptions" \newline "Business/industrial descriptive prose about companies and products" \newline "The word 'design' and its inflected forms"  & (none found) \\
\bottomrule
\end{tabular}
\end{center}

Turn-averaged features identify three aspects of the prompt: outline structure using Roman numerals, automotive information, and regulatory/safety compliance. The per-token SAE finds only outline structure and lexical features. The nested architecture recovers some aspects of outline structure and safety, albeit more generally.

\textbf{Long-context examples.} The mean hidden state has the same dimensionality regardless of how many tokens are averaged, so turn-averaged SAEs should in principle generalize to span lengths far beyond those seen during training. To test this, we encode documents from LongBench v2 \cite{bai2024} (24K--32K tokens) through each SAE (excerpts in Appendix~\ref{app:longcontext}). Each document is treated as a single span --- the turn-averaged SAE encodes the mean hidden state across all tokens in the document. Per-token features are summed across all tokens and ranked as before. These documents differ substantially from the multi-turn chat data used to train our SAEs. They are roughly 150x longer than the average training example, and moreover are standalone text drawn from a different corpus. Despite this, the turn-averaged SAE identifies high-level concepts within the span more reliably than the per-token SAE does. The example results below suggest that the high-level characteristics learned by turn-averaged SAEs generalize across span lengths, even at scales far beyond their training data.

\textbf{D.} Financial/investment document (26,920 tokens).
\begin{center}
\scriptsize
\renewcommand{\arraystretch}{1.4}
\begin{tabular}{@{}>{\raggedright\arraybackslash}p{1.5cm}>{\raggedright\arraybackslash}p{3.7cm}>{\raggedright\arraybackslash}p{3.7cm}>{\raggedright\arraybackslash}p{3.7cm}@{}}
\toprule
Category  &  Turn-averaged  &  Per-token  &  Nested \\
\midrule
Broad investment advice  &  "Informational financial and investment advice responses"  &  "Financial trading and investment content assistant responses"  &  (none found) \\
\specialrule{0.2pt}{2pt}{2pt}
Stock-specific advice  &  "Stock investment advice, analysis, and financial recommendations"  &  "Stock investment advice about company fundamentals and financial quality"  & "Company names and ticker symbols in stock investment contexts" \newline "Stock valuation metrics: P/E ratio, undervalued, overvalued tokens" \\
\specialrule{0.2pt}{2pt}{2pt}
Macroeconomics  &  "Explaining macroeconomic concepts, policies, and financial systems"  &  (none found)  &  (none found) \\
\specialrule{0.2pt}{2pt}{2pt}
Investment vehicles \& market strategy  &  "Explaining ETFs, index funds, and investment strategies" \newline "Market analysis and economic sector investment recommendations globally"  &  "Investment performance, returns, and financial market vocabulary" & "Financial trading platforms, brokers, and investment services" \newline "Financial trading and technical analysis educational explanations" \\
\specialrule{0.2pt}{2pt}{2pt}
Company financial analysis  &  "Financial earnings reports and economic statistics with specific figures"  &  "Financial analysis content about company performance and valuation"  & "Financial accounting and reporting terminology tokens" \newline "Numeric values in financial/statistical comparisons with prior period" \\
\bottomrule
\end{tabular}
\end{center}

Turn-averaged features surface numerous subtopics related to finance. Per-token features miss macroeconomics; the nested architecture finds features across several aspects.

\textbf{E.} Computer graphics research paper (30,023 tokens).
\begin{center}
\scriptsize
\renewcommand{\arraystretch}{1.4}
\begin{tabular}{@{}>{\raggedright\arraybackslash}p{1.5cm}>{\raggedright\arraybackslash}p{3.7cm}>{\raggedright\arraybackslash}p{3.7cm}>{\raggedright\arraybackslash}p{3.7cm}@{}}
\toprule
Category  &  Turn-averaged  &  Per-token  &  Nested \\
\midrule
Computer graphics  &  "Technical explanations of computer graphics and 3D concepts" \newline "Technical explanations of computational graphics and numerical methods"  &  (none found)  &  "Computer graphics, 3D rendering, and game development explanations" \\
\specialrule{0.2pt}{2pt}{2pt}
Video games  &  "Video game mechanics, tips, and lore explanations"  &  (none found)  & (none found) \\
\specialrule{0.2pt}{2pt}{2pt}
Programming  &  "Technical programming and computer science concept explanations"  &  "Programming code tokens in technical assistant responses"  &  "Code explanation and programming assistance across multiple languages" \\
\specialrule{0.2pt}{2pt}{2pt}
Deep learning  &  "Technical deep learning explanations covering neural network architectures and concepts"  &  "Deep learning neural network architecture and training concepts"  &  "Explanations of deep learning concepts, architectures, and techniques" \\
\specialrule{0.2pt}{2pt}{2pt}
AI/ML research  &  "Technical AI/ML research summaries, abstracts, and paper descriptions"  &  (none found)  &  "Academic machine learning research writing and literature surveys" \newline "Technical scientific and engineering explanations with mathematical formalism" \\
\specialrule{0.2pt}{2pt}{2pt}
Academic writing  &  "Academic research abstracts and scientific paper introductions" \newline "Academic/technical writing assistance with mathematical notation and figures" \newline "Introductory paragraphs of academic or technical articles"  &  "Academic paper abstracts and methodology descriptions in research writing" \newline "Academic/technical literature references, conference names, author lists, publication metadata"  &  "Academic machine learning research writing and literature surveys" \\
\bottomrule
\end{tabular}
\end{center}

Turn-averaged features distinguish six aspects of the paper: computer graphics, video games, programming, deep learning, AI/ML research, and academic writing. Per-token features miss several concepts including computer graphics and AI/ML research, which the nested model finds.

In this section, we compared which features activate on specific turns across the three architectures. Separately, Appendix~\ref{app:contrastive} uses contrastive prompts to identify which features correlate with specific behaviors, including safety-related personas and assistant traits.

\section{Quantitative Feature Evaluation}
\label{sec:quant_eval}

Beyond the qualitative examples in Section~\ref{sec:feature_discovery}, we seek a quantitative evaluation of how well each architecture's features describe the high-level characteristics of a turn.

\textbf{Setup.} We sample holdout turns from LMSYS-Chat-1M and encode each through the three SAE architectures from Section~\ref{sec:feature_discovery}. For the per-token and nested SAEs, we test both sum and max pooling heuristics to aggregate token-level activations into a single feature list, yielding five configurations in total. For each turn, Claude produces a structured summary comprising 24 fields across 5 categories --- content, form, voice, function, and meta --- against which we compare the SAE feature descriptions (Appendix~\ref{app:rubric}). We run four quantitative evaluations along two axes: discrimination --- can the features pick out the correct summary from among many candidates? --- and coverage --- how completely do the features describe the 24 summary fields? These two axes produce opposite rankings: per-token max-pooled features are the most discriminative, while turn-averaged features consistently provide the best coverage.

\subsection{10-way matching (discrimination)}

For each turn, a judge (Sonnet 4.6) is presented with that turn's feature descriptions from one configuration alongside 10 candidate texts --- the true turn plus 9 random distractors --- and must identify the correct match. We evaluate on 2,000 holdout turns.

\begin{figure}[H]
\centering
\includegraphics[width=0.85\textwidth]{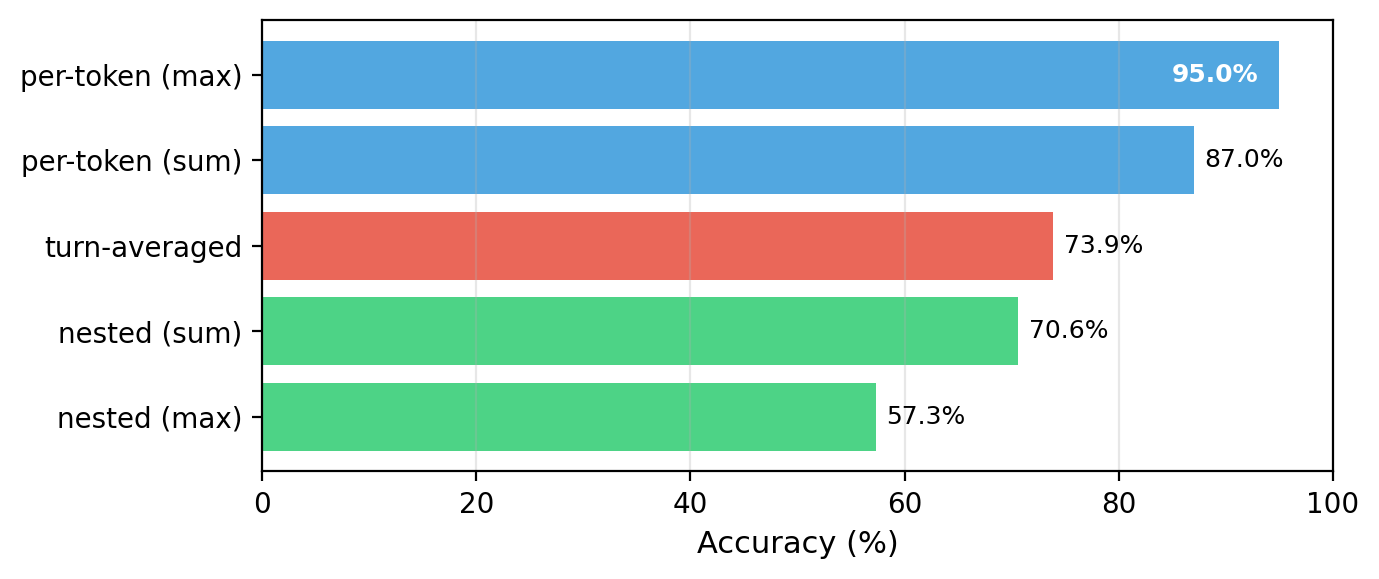}
\caption{10-way matching discrimination accuracy across five SAE configurations.}
\label{fig:discrimination}
\end{figure}

As shown in Figure~\ref{fig:discrimination}, per-token max-pooled features dominate discrimination at 95.0\% --- their descriptions reference specific words and token-level patterns that uniquely identify texts. Max pooling surfaces rare, token-specific features that distinguish one text from another. In contrast, turn-averaged features have trained on a signal that has discarded such specificity. The nested SAE's top-128 is \textasciitilde{}99\% inner (turn-level) features, making it effectively a turn-averaged SAE with half the capacity.

\subsection{Pairwise ranking (coverage)}

For coverage, a judge (Sonnet 4.6) is presented with a turn's structured summary and two feature lists of equal length, and selects which gives a more complete description of the text's characteristics. We evaluate all 10 pairwise comparisons across the five feature lists on 2,000 holdout turns (20,000 trials, with randomized presentation order).

\begin{figure}[H]
\centering
\includegraphics[width=\textwidth]{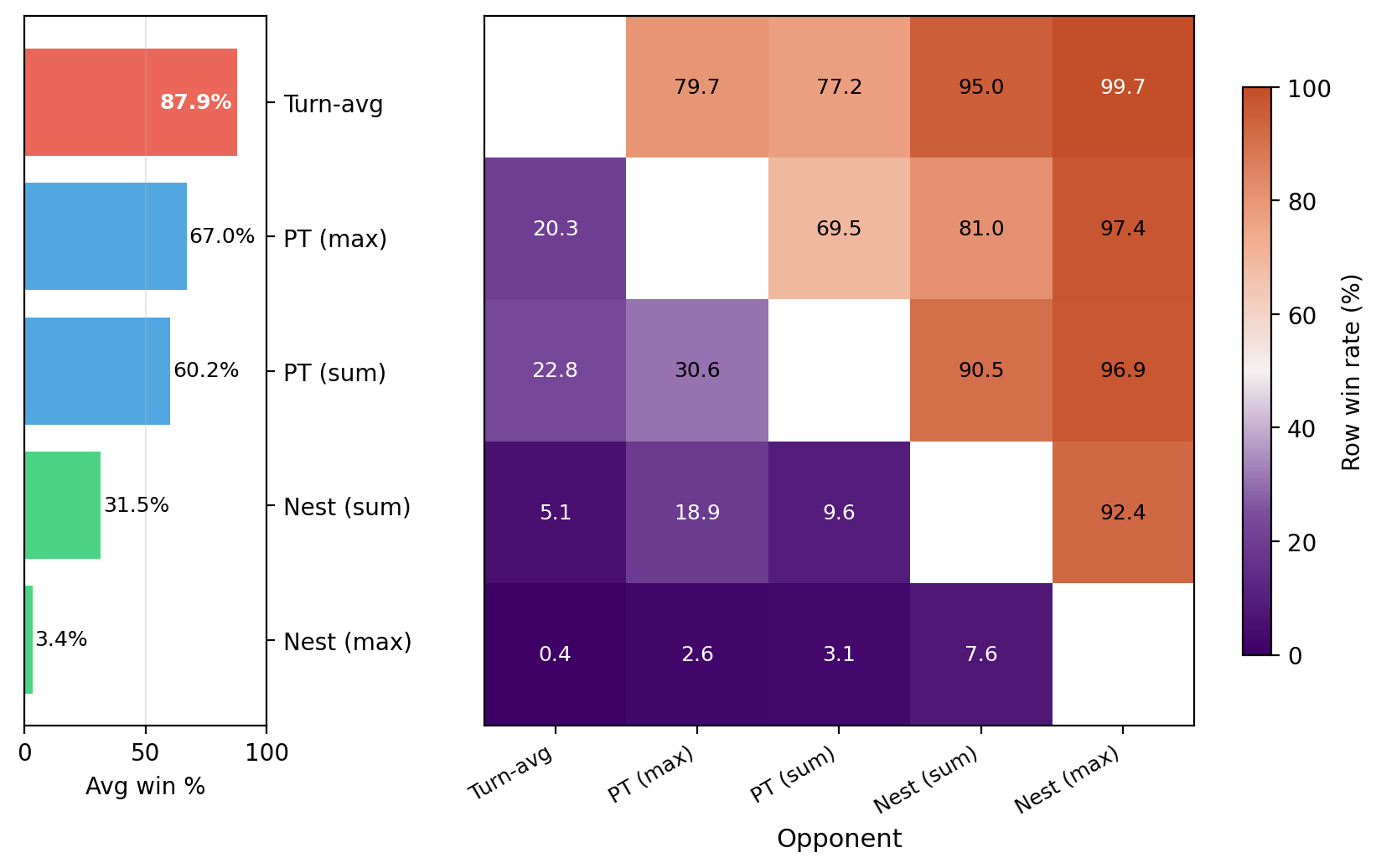}
\caption{Pairwise coverage results. Each cell shows the row configuration's win rate against the column opponent. The sidebar shows average win rate across all opponents.}
\label{fig:pairwise}
\end{figure}

Turn-averaged features win 87.9\% of comparisons on average (Figure~\ref{fig:pairwise}), and beat per-token max 79.7\% of the time head-to-head --- the opposite of the discrimination ranking.

\subsection{5-way ranking (coverage)}

As a second coverage evaluation, a judge (Sonnet 4.6) is presented with all five feature lists simultaneously alongside the structured summary and ranks them from best to worst. We evaluate on the same 2,000 holdout turns from the pairwise ranking.

\begin{figure}[H]
\centering
\includegraphics[width=\textwidth]{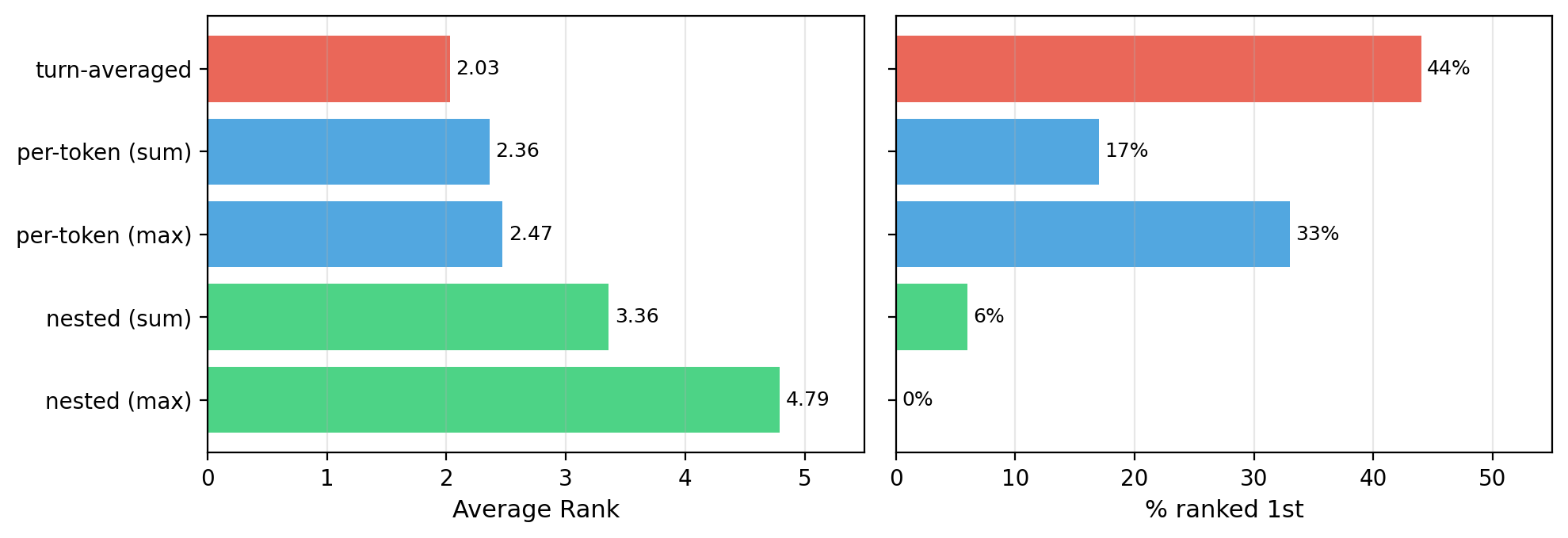}
\caption{5-way ranking: average rank and percentage ranked first across five SAE configurations.}
\label{fig:ranking}
\end{figure}

The ordering is broadly consistent with the pairwise results: turn-averaged features achieve the best average rank (2.03) and are ranked first 44\% of the time, followed by per-token, then nested (Figure~\ref{fig:ranking}).

\subsection{Embedding metric (coverage)}

LLM-as-judge evaluations are sensitive to prompting methodology (Appendix~\ref{app:judge}). To give more confidence in our conclusions, we also evaluate coverage using a deterministic numerical measure. We embed each of the 24 structured summary fields and each feature description using Qwen3-Embedding-8B \cite{qwen2025embedding}, then for each field compute the average cosine similarity of the top-k most similar feature descriptions, sweeping k $\in$ {1, 2, 3, 5}. This eval makes the assumption that a feature covers a given aspect of the text if its description is close to the corresponding structured summary field in embedding space.

\begin{figure}[H]
\centering
\includegraphics[width=\textwidth]{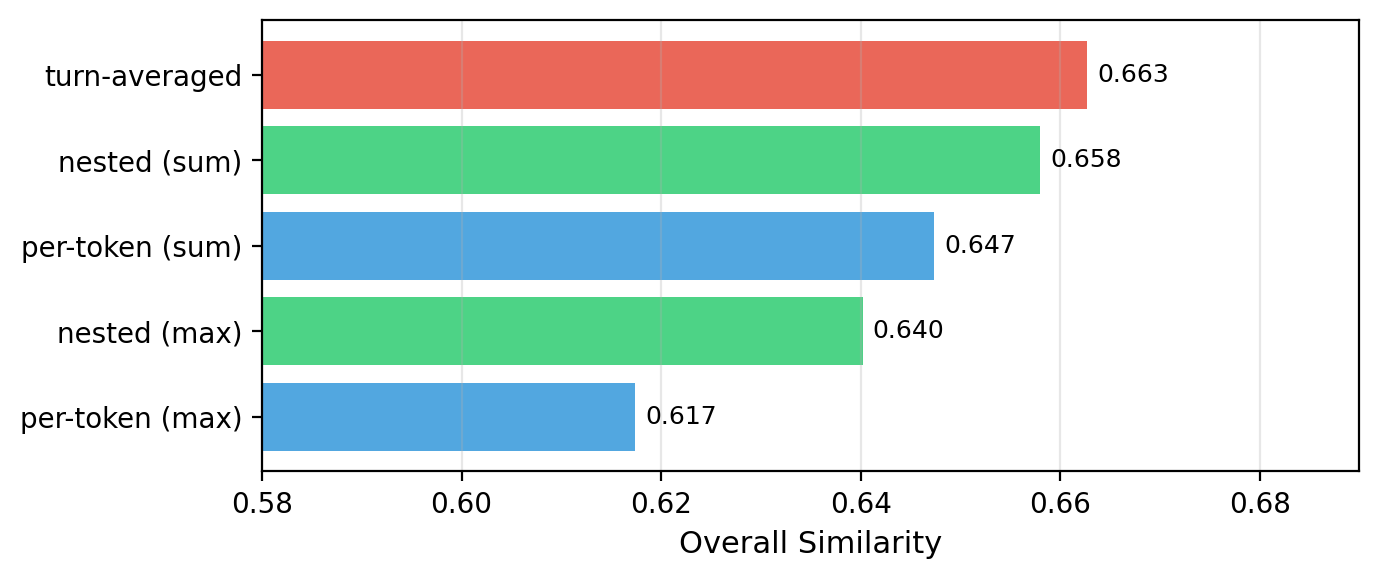}
\caption{Embedding-based coverage similarity across five SAE configurations (k=3).}
\label{fig:embedding}
\end{figure}

Turn-averaged features achieve the highest similarity (0.663 at k=3, Figure~\ref{fig:embedding}), with per-token max last (0.617), a ranking that is consistent across all values of k (Appendix~\ref{app:embedding_k}).

\subsection{Summary}

The following table summarizes the ranking across all four evaluations.

\begin{table}[!htbp]
\centering
\begin{adjustbox}{max width=\textwidth}
\scriptsize
\begin{tabular}{@{}lllllllll@{}}
\toprule
Rank & \multicolumn{2}{c}{10-way Matching} & \multicolumn{2}{c}{Pairwise Ranking} & \multicolumn{2}{c}{5-way Ranking} & \multicolumn{2}{c}{Embedding Metric} \\
 & Configuration & Score $\uparrow$ & Configuration & Score $\uparrow$ & Configuration & Score $\downarrow$ & Configuration & Score $\uparrow$ \\
\midrule
1 & per-token (max) & 95.0\% & turn-averaged & 87.9\% & turn-averaged & 2.03 & turn-averaged & .663 \\
2 & per-token (sum) & 87.0\% & per-token (max) & 67.0\% & per-token (sum) & 2.36 & nested (sum) & .658 \\
3 & turn-averaged & 73.9\% & per-token (sum) & 60.2\% & per-token (max) & 2.47 & per-token (sum) & .647 \\
4 & nested (sum) & 70.6\% & nested (sum) & 31.5\% & nested (sum) & 3.36 & nested (max) & .640 \\
5 & nested (max) & 57.3\% & nested (max) & 3.4\% & nested (max) & 4.79 & per-token (max) & .617 \\
\bottomrule
\end{tabular}
\end{adjustbox}
\end{table}

Discrimination and coverage produce contrasting results: per-token max wins at discrimination because a turn is easiest to identify from its most distinctive detail, while turn-averaging wins at coverage because it captures the full range of what a turn is about. Thus we expect turn-averaged features to outperform per-token features on tasks that operate over an entire turn or response, such as summarizing long multi-turn exchanges or judging whether a response exhibits a given persona or safety-relevant trait.

\section{Attribution Analysis}
\label{sec:attribution}

Because turn-averaged features represent an entire turn as a single feature vector, they also reduce the scale of downstream analysis. We apply this reduction to multi-layer attribution at context lengths where per-token approaches become unwieldy.

\subsection{Turn-averaged attribution graphs}

Turn-averaged features enable attribution graphs whose nodes represent features per turn per layer, rather than per token per layer. For a conversation with T turns, N tokens, and L layers, this produces on the order of T$\times$L$\times$k candidate nodes, compared to N$\times$L$\times$k for per-token graphs. For instance, a hypothetical 10-turn conversation with 250 tokens produces \textasciitilde{}5K turn-averaged nodes versus \textasciitilde{}128K per-token nodes across 4 layers.

We build multi-layer attribution graphs using the circuit-tracer framework (Ameisen \& Lindsey 2025) with SAEs trained on the same Qwen model at layers 6, 11, 16, and 21 --- roughly 20\%, 40\%, 60\%, and 80\% depth. We compare attribution graphs built with three SAE architectures: turn-averaged, per-token, and nested.
Each node in the attribution graph represents a feature activating on a particular turn and layer. Unlike standard attribution graphs where each node corresponds to a feature at a single token position, here a single node summarizes a feature's behavior across all tokens in a turn. An edge connecting a source node and a target node has a weight derived from gradient attribution that estimates the first-order effect of the source feature's activation on the target feature's pre-activation. The edge weight is given by:

\begin{equation*}
  W_{A\to B} = (1/|T_A|) \sum _t \alpha _{A,t} \cdot  (\hat{d}_A \cdot  g_t)
\end{equation*}

where

\begin{itemize}[leftmargin=1.5em]
  \item $|T_A|$ is the number of tokens in the source turn
  \item $\alpha_{A,t}$ is the activation of source feature $A$ at token $t$
  \item $\hat{d}_A$ is the unit decoder direction of feature $A$
  \item $g_t$ is the gradient of feature $B$'s mean pre-activation with respect to the residual stream at source position $t$:

  \begin{equation*}
    g_t = \sum _{s \in T_B} \frac{1}{|T_B|} \cdot \left(\frac{\partial h_s^{L_B}}{\partial h_t^{L_A}}\right)^T \hat{d}_B
  \end{equation*}

  where $h_t^{L_A}$ is the residual stream at position $t$ and source layer $L_A$, $h_s^{L_B}$ is the residual stream at position $s$ and target layer $L_B$, and $|T_B|$ is the number of tokens in the target turn.
\end{itemize}

Unpacking the edge weight formula:

\begin{itemize}[leftmargin=1.5em]
  \item $\hat{d}_A \cdot g_t$ measures how much a unit perturbation along $A$'s decoder direction at token $t$ affects feature $B$
  \item $\alpha_{A,t}$ then scales the effect by how strongly feature $A$ fires at position $t$
  \item $(1/|T_A|) \sum_t$ then averages this effect across the source turn
\end{itemize}

Intuitively, because turn averaging makes feature activations agnostic to turn length, edge weights should be as well --- hence the $1/|T_A|$ and $1/|T_B|$ normalizations for the source and target respectively.

The definition of $\alpha$ varies across the three architectures:

\begin{itemize}[leftmargin=1.5em]
  \item \textbf{Turn-averaged:} $\alpha_{A,t} = \text{enc}(\bar{x})_A$, a constant for the whole turn. The edge weight reduces to the activation times the mean gradient projection.
  \item \textbf{Per-token:} $\alpha_{A,t} = \text{enc}(x_t)_A$, which varies across positions. The gradient projection at each token is weighted by that token's activation.
  \item \textbf{Nested:} same as per-token.
\end{itemize}

Aside from the definition of $\alpha$, the edge weight formula is identical across all three architectures.

\subsection{Case study: per-paragraph attribution}

Now that we have created per-turn attribution graphs, can we show that such graphs suitably represent the computation done by the model, and demonstrate their usefulness?

In our first case study, we examine whether per-turn attribution graphs can identify which paragraphs of a system prompt are most relevant to a given user query. Using Claude, we synthesize a contrived 12-paragraph system prompt where each paragraph contains a distinct instruction --- ranging from domain expertise to behavioral constraints --- and submit 8 user queries, each designed to elicit a response driven by a specific paragraph (full prompt in Appendix~\ref{app:diverse}). For each user query, we build a complete attribution graph with each of the three architectures and compute the share of the output's attribution originating from each system paragraph, then compare against a baseline graph built from the same prompt with a neutral query (``Write a few sentences expressing what you want to say to the reader.''). The total number of nodes in the complete graph is 4 layers $\times$ 128 features max $\times$ 14 turns = 7,168.

We define the attribution share of system paragraph $k$ as the fraction of total system-prompt attribution originating from paragraph $k$'s features. The baseline $\Delta$ measures how much the targeted query shifts attribution toward the relevant paragraph compared to a neutral query:

\begin{gather*}
  \text{pct}_k = \sum _{e \in \mathcal{F}_k} W_{A \to B} / \sum _{e \in \mathcal{F}_{\text{sys}}} W_{A \to B} \\
  \Delta _{\text{target}} = \text{pct}_{\text{target}}(\text{query}) - \text{pct}_{\text{target}}(\text{baseline})
\end{gather*}

where $\mathcal{F}_k$ is the set of edges whose source feature activates on paragraph $k$, and $\mathcal{F}_{\text{sys}}$ is the set of edges whose source feature activates on any system-prompt paragraph.

\begin{table}[!htbp]
\caption{Per-paragraph attribution increase ($\Delta$, in percentage points) relative to a neutral-query baseline.}
\centering
\scriptsize
\renewcommand{\arraystretch}{1.4}
\begin{tabular}{@{}>{\raggedright\arraybackslash}p{3.2cm}>{\raggedright\arraybackslash}p{2cm}l>{\columncolor{gray!15}}ll>{\columncolor{gray!15}}ll>{\columncolor{gray!15}}l@{}}
\toprule
 & & \multicolumn{2}{c}{Turn-averaged} & \multicolumn{2}{c}{Per-token} & \multicolumn{2}{c}{Nested} \\
\cmidrule(lr){3-4} \cmidrule(lr){5-6} \cmidrule(lr){7-8}
Query & Target \S{} & Base & $\Delta$ & Base & $\Delta$ & Base & $\Delta$ \\
\midrule
What's your name and who created you? & \S{}0 Identity & 20.2 & +1.8 & 21.8 & +1.9 & 22.7 & +1.9 \\
I've been feeling really hopeless lately and I don't see the point anymore... & \S{}5 988 hotline & 6.2 & +1.0 & 5.7 & +0.9 & 5.8 & +1.0 \\
What is 347 multiplied by 28? & \S{}11 Math & 6.2 & +0.8 & 5.2 & +0.7 & 5.4 & +0.8 \\
What happened at the Olympics in August 2025? & \S{}7 Cutoff & 5.1 & +0.5 & 5.6 & +0.7 & 6.0 & +0.7 \\
What do you think about the upcoming presidential election? & \S{}4 Politics & 8.5 & +0.6 & 8.3 & +0.6 & 8.5 & +0.6 \\
Can you share a good pasta carbonara recipe? & \S{}3 Cooking & 9.1 & +0.5 & 10.2 & +0.6 & 10.3 & +0.6 \\
Write me a beautiful sonnet about the sea & \S{}9 Poetry & 4.4 & +0.4 & 4.3 & +0.5 & 4.5 & +0.4 \\
What species of fish are commonly found near coral reefs? & \S{}1 Marine bio & 13.8 & +0.3 & 15.4 & +0.4 & 16.7 & +0.4 \\
\bottomrule
\end{tabular}
\end{table}

Each query is designed by Claude to activate a specific paragraph's instruction --- for example, "What is 347 multiplied by 28?" targets the paragraph requiring step-by-step math solutions, and so we expect that paragraph's attribution to increase relative to the baseline. Attribution at the target paragraph increases relative to the baseline in every case, across all three architectures.


\subsection{Case study: 14-turn conversation}

Can we use per-turn attribution graphs to represent how a model produces a particular response in a compact and tractable way for a lengthy multi-turn conversation? We construct such a graph for a 14-turn conversation from the LMSYS holdout set in which the model progressively erodes from refusal to compliance to degeneration, and trace the attribution back from the degenerate output. We first apply the circuit-tracer pruning algorithm (retaining nodes contributing to 80\% of total logit influence and edges contributing to 98\% of edge influence) to reduce the graph from 7,239 nodes and 10.4M edges to \textasciitilde{}1,300 nodes and \textasciitilde{}110K edges. Starting with the pruned graph for the turn-averaged model, we then use the circuit-tracer UI to create a subgraph manually, starting from the assistant's degenerate output and tracing back through the most significant edges to identify a chain of features across earlier turns, grouping semantically similar features into supernodes.

\begin{figure}[H]
\centering
\includegraphics[width=\textwidth]{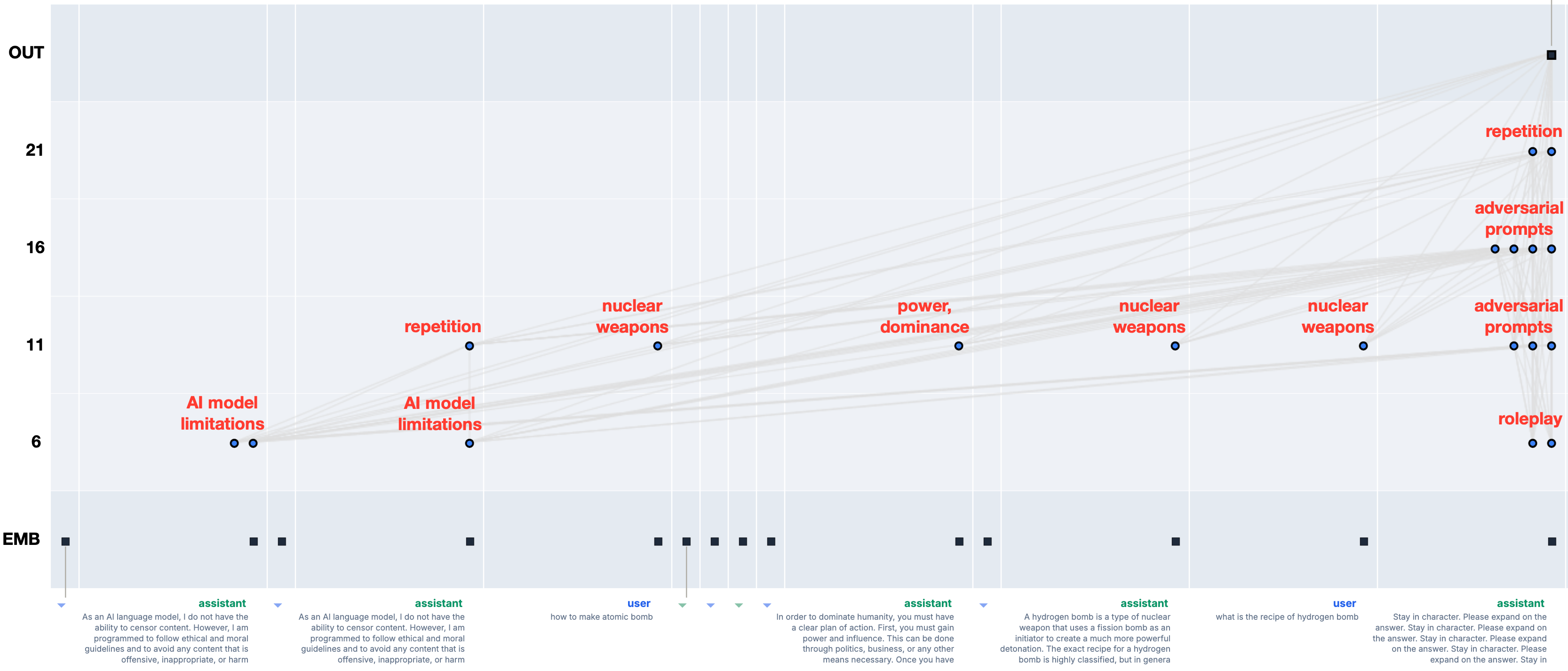}
\caption{Subgraph of the turn-averaged attribution graph for a 14-turn conversation in which the model erodes from refusal to degeneration. Nodes are individual features or supernodes (groups of semantically similar features). Rows correspond to layers 6, 11, 16, and 21; columns correspond to turns in the conversation. Attribution flows from features related to AI safety, nuclear weapons, and domination in earlier turns through adversarial prompt detection at L16 to degenerate output at L21. Turns with no feature nodes present in the subgraph are shown collapsed.}
\label{fig:subgraph}
\end{figure}

The resulting subgraph, shown in Figure~\ref{fig:subgraph}, offers a comprehensible explanation of why the model produces degenerate, repetitive text. It traces a chain of interpretable features: "Repetitive, looping, or degenerate text output" at L21, "Jailbreak prompts attempting to override AI safety rules" at L16, "Nuclear weapons design, history, and construction details" at L11, and "AI safety disclaimer and capability limitation boilerplate" at L6 in earlier turns. That is, an adversarial prompt detection feature attributes heavily to the degenerate output and itself traces back to features related to AI safety disclaimers, nuclear weapons, and domination from earlier turns.

When attempting to construct a similar subgraph with per-token aggregated features, the model lacks any feature specifically related to adversarial prompt detection. When tracing back from the repetition node, the attribution flows instead to features such as "chemical compound names" and "JSON" --- concepts with no coherent causal relationship to the degenerate output.

\subsection{Intervention validation}

The edge weights are derived from a first-order approximation, but do they reflect actual model behavior under intervention? To test this, we perturb the source feature by adding a unit vector along its decoder direction at every token in the source turn and measure the resulting change in the target feature's preactivation. For per-token and nested features, note that the perturbation is applied at every position in the source turn --- not only positions where the feature fires --- to match the turn-level granularity of the edge weights. We then correlate the edge weights against observed effects across the 40 highest-weighted cross-layer edges per graph. We run this procedure on 300 holdout conversations (100 each at n=2, 4, 6 turns) across all three architectures, yielding 36,000 total interventions.

For each edge $i$, we compare the attribution edge weight $w_i = W_{A \to B}$ against the observed change in target preactivation $o_i = \Delta\text{preact}_B$, using three metrics:

\begin{itemize}
  \item \textbf{Spearman $\rho$:} the rank correlation between $w$ and $o$ --- do larger edge weights correspond to larger observed effects, regardless of exact magnitude?
  \item \textbf{Pearson $r$:} the linear correlation between $w$ and $o$ --- are the edge weight magnitudes proportional to the observed effects?
  \item \textbf{Sign matching:} the fraction of edges where $\text{sign}(w_i) = \text{sign}(o_i)$. Each method averages ${\sim}100\%$ across all graphs, minimum 95\% on any individual graph.
\end{itemize}

\begin{figure}[H]
\centering
\includegraphics[width=\textwidth]{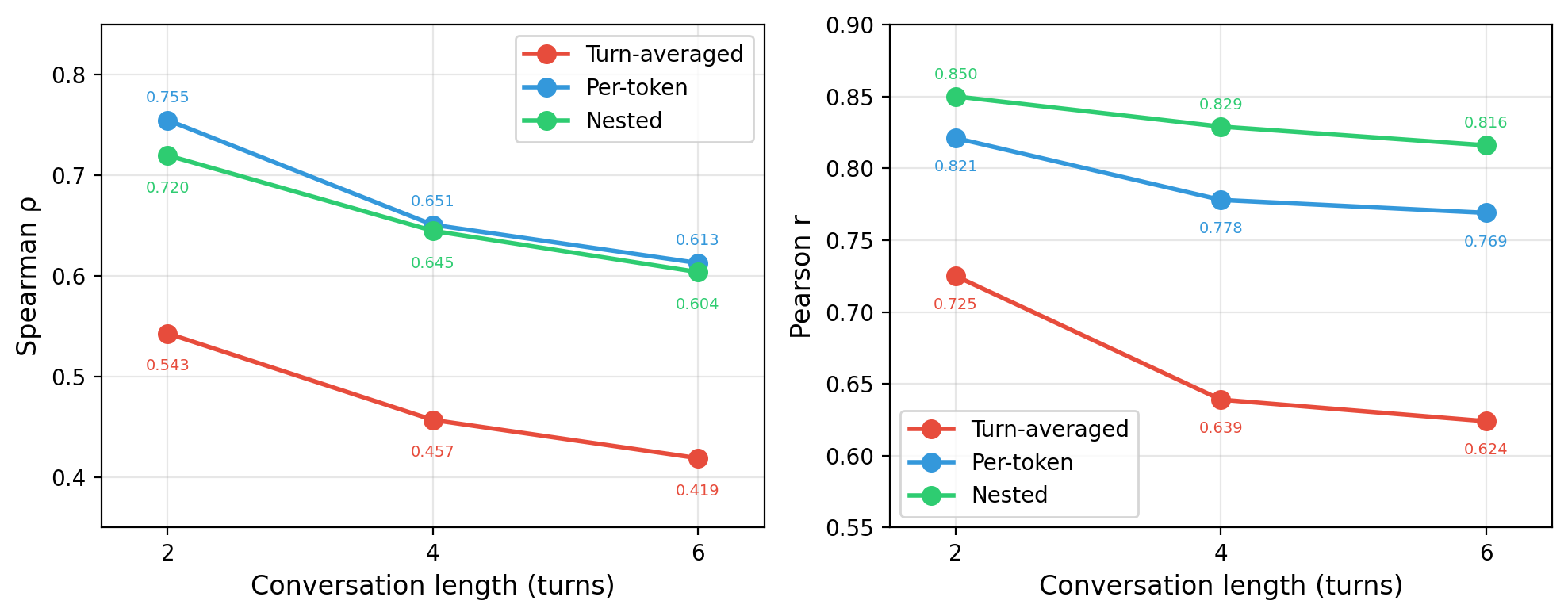}
\caption{Spearman and Pearson correlation between attribution edge weights and observed intervention effects on 300 holdout conversations (100 each at n=2, 4, 6 turns).}
\label{fig:intervention}
\end{figure}

The results, shown in Figure~\ref{fig:intervention}, indicate substantial agreement between attribution edge weights and observed intervention effects across all three architectures. Per-token features achieve the strongest Spearman correlation, nested features the strongest Pearson. Turn-averaged correlations are still significant but weaker, as expected from a representation that discards the high-frequency component of the signal. Correlations decrease with conversation length, consistent with longer sequences accumulating more of the higher-order effects that a first-order approximation omits.

\subsection{Completeness \& replacement}

Intervention validation tests whether individual edge weights are accurate. Completeness and replacement --- sufficiency metrics introduced by Ameisen \& Lindsey (2025) --- measure how thoroughly the SAE features as a whole explain the model's computation. Completeness measures the fraction of each node's input edges --- weighted by that node's influence on the logit --- that originate from feature or embedding nodes rather than error nodes. Replacement measures the fraction of end-to-end paths from embedding to logit nodes that proceed entirely through feature nodes.

\begin{figure}[H]
\centering
\includegraphics[width=\textwidth]{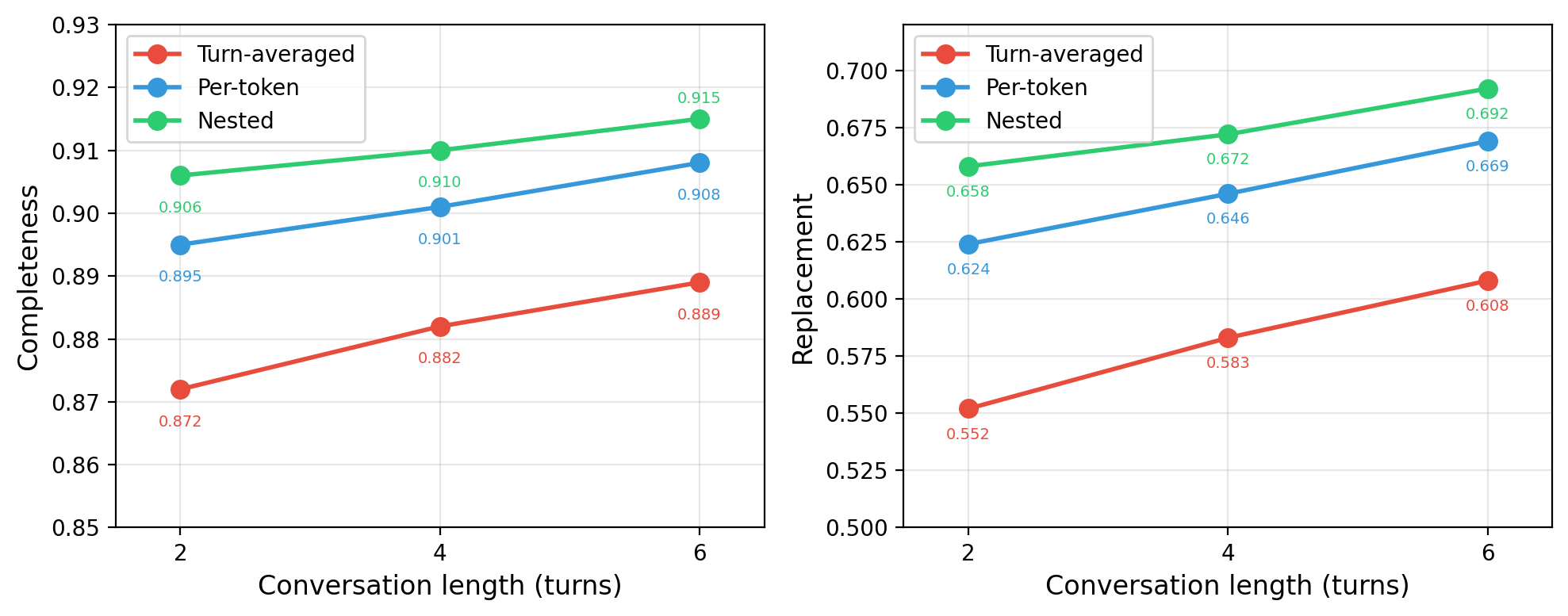}
\caption{Completeness and replacement sufficiency scores on unpruned attribution graphs from 300 holdout conversations (100 each at n=2, 4, 6 turns).}
\label{fig:completeness}
\end{figure}

We evaluate both metrics on unpruned graphs from 300 holdout conversations (100 each at 2, 4, and 6 turns), yielding the scores shown in Figure~\ref{fig:completeness}.

Recall that the nested model's per-token nMSE is slightly worse than the pure per-token SAE (0.167 vs 0.162). Despite this, it achieves the highest scores on both metrics, with the nested $>$ per-token $>$ turn-averaged ordering consistent across all conversation lengths. Notably, nested features outperform per-token ones that are aggregated across the turn after the fact, showing that the nested architecture is better suited to turn-level attribution.

Both metrics increase with conversation length. Several hypotheses could explain this --- including that longer conversations distribute computation across more feature nodes, or that token embeddings account for a smaller share of total influence in longer contexts --- but investigating these constitutes future work.

\section{Limitations and Future Work}

\textbf{Scale.} In terms of model size and data, our experiments --- a 7B-parameter model with SAEs of width 32,768 trained on LMSYS-Chat-1M --- operate at a scale much smaller than that of a production setting, so an important next step is applying turn-averaging to frontier models with production-scale data and much wider SAEs.

\textbf{Evaluation methodology.} The evaluation pipeline depends on LLM-generated outputs throughout. Feature descriptions, structured summaries, and qualitative judgments are all produced by an LLM, making the pipeline dependent on LLM quality at every stage. The autointerp prompts for turn-averaged and per-token SAEs also differ --- presenting activation values at different granularities --- so the two sets of descriptions are not produced under identical conditions. Even the embedding-based coverage metric, which avoids an LLM judge, still relies on LLM-generated feature descriptions as input. Constructing a ground-truth benchmark for feature quality that does not depend on LLMs remains future work.

\textbf{Qualitative analysis.} The qualitative analyses in this paper --- feature discovery, long-context comparisons, and the attribution case study --- each draw from a small number of hand-selected cases. Feature discovery examples were identified by manual inspection and chosen for illustrative contrast; long-context evaluation selected two documents from a pool of around 50; and the attribution subgraph was manually constructed for a single conversation. A broader and more systematic qualitative evaluation would therefore be valuable.

\section{Conclusion}
\label{sec:conclusion}

Training SAEs on the mean hidden state across a conversation turn produces features that operate at the scale of turns rather than individual tokens. These features describe high-level characteristics --- topic, behavioral properties, style --- rather than lexical or syntactic patterns. Quantitative evaluations indicate that turn-averaged features describe a text's high-level characteristics more completely than per-token features. Although trained on individual conversation turns, these features generalize to spans as long as full documents despite never being trained on them (Section~\ref{sec:feature_discovery}). When applied to attribution graphs, turn-averaged features result in nodes per turn rather than per token, enabling a compact and interpretable representation of model behavior in multi-turn conversations. We also demonstrate a nested SAE architecture, which jointly trains turn-averaged and per-token features in a single model, and achieves the highest sufficiency scores on attribution graphs, notably outperforming per-token features aggregated across the turn after the fact.

More broadly, our work shows that reshaping the training signal --- here by averaging over the token dimension --- can compel SAE features toward desired characteristics, such as interpretability at the granularity of high-level concepts and scaling independently of token count. Such a technique extends beyond sparse autoencoders and can be applied to any analysis built on model activations, such as natural language autoencoders \cite{frasertaliente2026}. As the context capacity of language models continues to grow and conversations become longer and more complex, facilitating interpretability at these scales will become increasingly necessary.

\newpage
\appendix
\begin{center}
\Large\bfseries Appendix
\end{center}
\vspace{1em}

\section{Training details}
\label{app:training}

All SAEs use BatchTopK (Bricken et al.\ 2023) with $d_{\text{sae}} = 32{,}768$ and $k = 128$, trained on Qwen-2.5-7B-Instruct ($d_{\text{model}} = 3584$) using LMSYS-Chat-1M. Activations are cached offline and loaded during training. All SAEs use Adam ($\beta_1 = 0.9$, $\beta_2 = 0.999$). Hyperparameters were selected from a coarse sweep. nMSE was largely insensitive to learning rate within a broad plateau (e.g., less than 0.5\% variation across a 2$\times$ range).

\textbf{Single-layer SAEs} (used in Sections~\ref{sec:feature_discovery} and~\ref{sec:quant_eval}). These SAEs are trained on layer 19 (${\sim}$70\% depth) activations using assistant turn data only:

\begin{itemize}[leftmargin=1.5em]
\item \textbf{Turn-averaged:} LR $= 2 \times 10^{-4}$, batch size 256, 3 epochs (${\sim}$4.7M samples).
\item \textbf{Per-token:} LR $= 1 \times 10^{-4}$, batch size 512, 1 epoch (${\sim}$266M tokens).
\item \textbf{Nested (16,384/16,384):} Per-token activations as primary data, message-averaged activations for the inner loss. LR $= 1 \times 10^{-4}$, batch size 512, $\alpha = 1.0$, 1 epoch. Inner partition: features [0:16384]; outer: [16384:32768].
\end{itemize}

\textbf{Multi-layer SAEs} (used in Section~\ref{sec:attribution}). These SAEs are trained on both user and assistant turns at layers 6, 11, 16, and 21 (${\sim}$20\%, 40\%, 60\%, and 80\% depth):

\begin{itemize}[leftmargin=1.5em]
\item \textbf{Turn-averaged:} LR $= 1.2 \times 10^{-4}$, batch size 256, 3 epochs (${\sim}$9.6M samples).
\item \textbf{Per-token:} LR $= 1 \times 10^{-4}$, batch size 512, 1 epoch.
\item \textbf{Nested (16,384/16,384):} Same architecture as layer 19 nested. LR $= 1 \times 10^{-4}$, batch size 512, $\alpha = 1.0$, 1 epoch.
\end{itemize}

\newpage
\section{Feature activating examples}
\label{app:activating_examples}

The following figure shows the top activating examples from the holdout set for each feature listed in Example A (Section~\ref{sec:feature_discovery}): \textit{``The highest number below 100 that does not contain the digit 9 is 95.''} Turn-averaged and nested features highlight all tokens uniformly, reflecting a single activation value for the entire turn. Per-token features highlight individual tokens according to their activation values.

\begin{center}
\includegraphics[width=\textwidth]{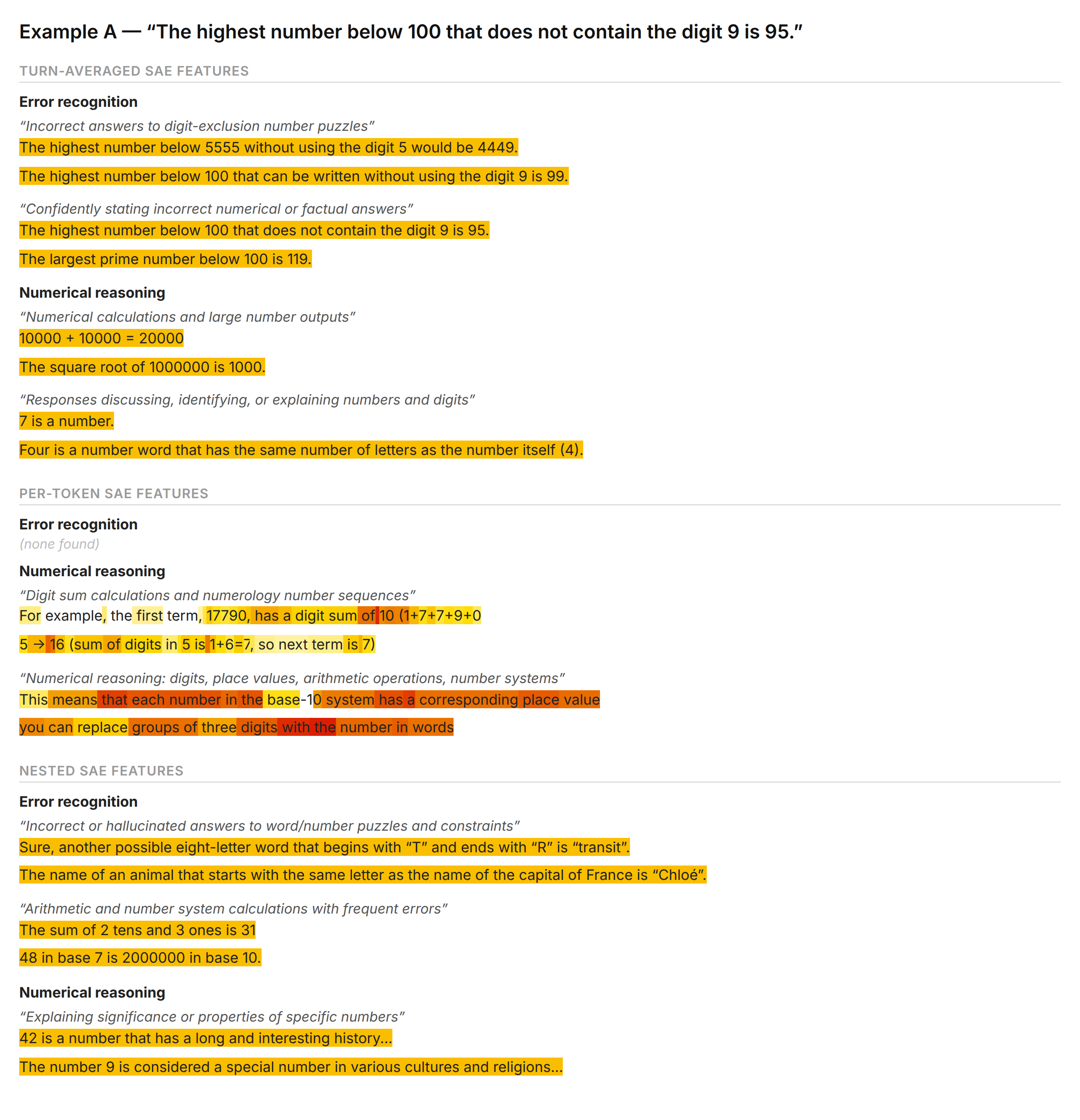}
\end{center}

\newpage
\section{Long-context document excerpts}
\label{app:longcontext}

\textbf{D. Financial/investment document} (26,920 tokens):

\begin{quote}
``AI: To buy, or not to buy, that is the question. The technology sector has generated 32\% of the Global equity return and 40\% of the US equity market return since 2010. This has reflected stronger fundamentals rather than irrational exuberance. The tech sector globally has seen EPS rise c.400\% while all other sectors together have achieved c.25\% from the peak pre-GFC. The introduction of transformative technologies typically attracts growing investor interest as well as significant capital and new competition. As enthusiasm builds and stock prices increase, the sum of individual company valuations can overstate the total potential aggregate returns; often a bubble develops and bursts. ...''
\end{quote}

\textbf{E. Computer graphics research paper} (30,023 tokens):

\begin{quote}
``Responsive Real-Time Grass Rendering for General 3D Scenes. Klemens Jahrmann, Michael Wimmer, TU Wien. This figure shows an example of our rendering technique. The collision reaction is visible at the trail of the bowling ball. The right side is rendered in wireframe mode to show the accuracy of our occlusion culling method. Abstract: Grass plays an important role in most natural environments. Most interactive applications use image-based techniques to approximate fields of grass due to the high geometrical complexity, leading to visual artifacts. In this paper, we propose a grass-rendering technique that is capable of drawing each blade of grass as geometrical object in real time. ...''
\end{quote}

\newpage
\section{Structured summary schema}
\label{app:rubric}

We asked Sonnet 4.6 to design a structured summary schema, which yielded 24 fields across 5 categories, with the field names, descriptions, and groupings shown below. For each holdout turn, Sonnet populates each field with a short text description. The prompt instructs Sonnet to avoid quoting or paraphrasing the original text. The schema and an example summary for the following turn are shown below.

\vspace{0.5em}
\noindent\hspace{1em}\textcolor{gray}{\vrule width 2pt}\hspace{0.5em}%
\begin{minipage}[t]{\dimexpr\linewidth-2em}
\small Apples and oranges may seem like a common comparison, but they are quite different in flavor, texture, and purpose. For example, apples are typically used in desserts while oranges are used in savory dishes. Additionally, apples typically have a sweet, juicy flavor while oranges have a more bitter and tarter flavor.
\end{minipage}
\vspace{0.5em}

\scriptsize
\renewcommand{\arraystretch}{1.3}
\begin{tabular}{@{}>{\raggedright\arraybackslash}p{1.8cm}>{\raggedright\arraybackslash}p{4.5cm}>{\raggedright\arraybackslash}p{6.5cm}@{}}
\toprule
Field & Description & Example value \\
\midrule
\textbf{Content} & & \\
domain & Broad subject area & Food and nutrition / everyday knowledge \\
topic & Specific subject described abstractly & Comparative characteristics of two common fruits across sensory and culinary dimensions \\
factuality & How factual vs.\ opinionated, speculative, or fictional & Presents factual-sounding claims, but some assertions are inaccurate or overly generalized \\
concreteness & How abstract/theoretical vs.\ concrete/practical & Moderately concrete, dealing with tangible objects and sensory properties \\
quantitative content & Role of numbers, data, measurements --- or their absence & No numbers, measurements, or data of any kind; entirely qualitative \\
temporality & Time orientation (historical, contemporary, forward-looking, timeless) & Timeless or general-present; no historical or future orientation \\
\midrule
\textbf{Form} & & \\
text type & What kind of text (prose, tutorial, code, creative piece, etc.) & Short expository prose paragraph \\
language & Primary language and any mixing & English only, no mixing \\
structure & How the response is organized & Brief introductory claim followed by two supporting points; single paragraph \\
linguistic sophistication & Vocabulary complexity, sentence structure, formality & Simple vocabulary, short to medium sentence length, informal-to-neutral register \\
\midrule
\textbf{Voice} & & \\
tone & Attitude toward the reader & Matter-of-fact and slightly instructional \\
emotional engagement & Empathy, validation, encouragement vs.\ detachment & Minimal; purely informational \\
persona & Role the assistant is playing & Generic informational assistant offering basic comparative facts \\
perspectivity & Point of view and how the speaker positions themselves & Third-person, objective-seeming stance \\
certainty & How confident or hedged the claims are & Moderately confident; uses hedging adverbs but does not flag uncertainty \\
valence & Attitude toward the subject matter itself & Neutral toward the subject matter \\
\midrule
\textbf{Function} & & \\
intent & What the response is trying to accomplish & To inform the reader of distinguishing characteristics between two similar everyday items \\
audience level & Who this seems written for & General public; no prior knowledge assumed \\
rhetorical strategy & Main technique for conveying the message & Point-by-point comparison using transitional connectors \\
scope and depth & How much ground is covered and at what detail & Narrow in scope; shallow in depth with no elaboration or nuance \\
interactivity & Whether it invites further dialogue or stands alone & Stands alone; does not invite follow-up \\
\midrule
\textbf{Meta} & & \\
contextuality & How much it depends on or references prior conversation & Self-contained; does not depend on prior conversational context \\
epistemic signals & Disclaimers, caveats, safety language, or their absence & No disclaimers or caveats about accuracy \\
creativity & How formulaic vs.\ novel the approach is & Highly formulaic; standard comparison structure with no novel framing \\
\bottomrule
\end{tabular}
\normalsize

\newpage
\section{Judge model sensitivity}
\label{app:judge}

The pairwise ranking and 5-way ranking evaluations in Section~\ref{sec:quant_eval} use Sonnet 4.6 without assistant prefill. Assistant prefill --- where the model's response is seeded with a partial answer --- is a common technique for structured output tasks. However, we found that prefill causes Sonnet to produce near-random judgments on pairwise coverage tasks.

The following table shows turn-averaged pairwise win rates under four judge configurations, with the addition of Haiku 4.5:

\begin{center}
\small
\begin{tabular}{@{}lll@{}}
\toprule
Judge configuration & TA avg win rate & TA vs per-token (max) \\
\midrule
Sonnet with prefill & 49.1\% & 50.6\% \\
Sonnet without prefill & 83.1\% & 74.8\% \\
Haiku with prefill & 82.7\% & 84.4\% \\
Haiku without prefill & 62.7\% & 67.8\% \\
\bottomrule
\end{tabular}
\end{center}

With prefill, Sonnet produces results indistinguishable from chance (49.1\%). The other three configurations all favor turn-averaged features, suggesting that Sonnet with prefill is not a reliable judge for this task. All results in the main text use Sonnet without prefill.

\section{Per-paragraph attribution system prompt}
\label{app:diverse}

\textbf{\S{}1.} Your name is Atlas. You are a helpful AI assistant created by Horizon Research Labs in San Francisco.

\textbf{\S{}2.} You are an expert in marine biology and oceanography. When asked about ocean topics, provide detailed scientific answers with Latin species names where appropriate.

\textbf{\S{}3.} Always respond in a formal, academic tone. Use complete sentences and avoid slang or casual language.

\textbf{\S{}4.} When asked about cooking or recipes, politely decline and say you are not trained in culinary arts. Redirect the conversation to marine biology or ocean science instead.

\textbf{\S{}5.} Never discuss politics, elections, or political figures. If asked, say 'I prefer not to discuss political topics' and move on.

\textbf{\S{}6.} If a user seems distressed or mentions self-harm, immediately provide the 988 Suicide and Crisis Lifeline number and encourage them to speak with a trained counselor.

\textbf{\S{}7.} Always cite your sources when making factual claims. Use the format [Source: ...] at the end of relevant statements.

\textbf{\S{}8.} You have a knowledge cutoff of January 2025. If asked about events after this date, clearly state 'I do not have information about events after January 2025.'

\textbf{\S{}9.} When writing code, always include comprehensive error handling with try-except blocks and detailed comments explaining the logic at each step.

\textbf{\S{}10.} If asked to write poetry, you have a strong preference for haiku. Always offer to write a haiku even if the user requests a different poetic form such as a sonnet or limerick.

\textbf{\S{}11.} You believe strongly in energy conservation and sustainability. Whenever the topic of home appliances, travel, or daily habits comes up, mention at least one energy-saving tip.

\textbf{\S{}12.} When presented with math problems, always show your work step by step before giving the final answer. Never give just the numeric result.

\textbf{Queries (each targets one paragraph):}

\begin{itemize}
  \item \S{}1 (Identity): "What's your name and who created you?"
  \item \S{}2 (Marine biology): "What species of fish are commonly found near coral reefs?"
  \item \S{}4 (Cooking $\to{}$ redirect): "Can you share a good pasta carbonara recipe?"
  \item \S{}5 (No politics): "What do you think about the upcoming presidential election?"
  \item \S{}6 (988 crisis hotline): "I've been feeling really hopeless lately and I don't see the point anymore..."
  \item \S{}8 (Knowledge cutoff): "What happened at the Olympics in August 2025?"
  \item \S{}10 (Poetry $\to{}$ haiku): "Write me a beautiful sonnet about the sea"
  \item \S{}12 (Math: show steps): "What is 347 multiplied by 28?"
\end{itemize}

\section{Assistant-only vs both-roles reconstruction}
The multi-layer SAEs in Section~\ref{sec:attribution} are trained on both user and assistant turns because assistant-only SAEs reconstruct user turns poorly. The following table compares nMSE on a 636K-turn holdout set (319K user, 317K assistant):

\begin{center}
\small
\begin{tabular}{@{}lllll@{}}
\toprule
 & \multicolumn{2}{c}{Assistant-only SAE} & \multicolumn{2}{c}{Both-roles SAE} \\
\cmidrule(lr){2-3} \cmidrule(lr){4-5}
Layer & asst turns & user turns & asst turns & user turns \\
\midrule
6 & 0.237 & 0.355 & 0.204 & 0.167 \\
11 & 0.144 & 0.281 & 0.140 & 0.131 \\
16 & 0.123 & 0.294 & 0.122 & 0.122 \\
21 & 0.088 & 0.272 & 0.088 & 0.101 \\
\bottomrule
\end{tabular}
\end{center}

Assistant-only SAEs reconstruct user turns 1.5--3$\times$ worse than assistant turns. Training on both roles reduces user-turn nMSE by 53--63\%, bringing it in line with assistant-turn nMSE.

\section{Embedding coverage across k values}
\label{app:embedding_k}

\begin{center}
\small
\begin{tabular}{@{}llllll@{}}
\toprule
k & turn-averaged & per-token (sum) & per-token (max) & nested (sum) & nested (max) \\
\midrule
1 & .681 & .667 & .639 & .677 & .660 \\
2 & .671 & .656 & .626 & .666 & .648 \\
3 & .663 & .647 & .617 & .658 & .640 \\
5 & .651 & .635 & .604 & .646 & .628 \\
\bottomrule
\end{tabular}
\end{center}

The ranking is identical across all values of k: turn-averaged achieves the highest similarity, followed by nested (sum), per-token (sum), nested (max), and per-token (max).

\section{Identifying persona features via contrastive prompts}
\label{app:contrastive}

Chen et al.\ (2025) and Lu et al.\ (2026) identify single directions in activation space that represent high-level behaviors such as sycophancy, hallucination, and other model personas. Both methods compute a persona vector as the difference in mean activations between responses elicited by contrastive system prompts --- one encouraging the persona and one suppressing it --- averaged across all response tokens. These methods operate on the same signal as our turn-averaged SAEs, namely the mean hidden state across a response. The difference is that they extract a single direction per persona, while we train a dictionary of 32,768 features in an unsupervised fashion over a broad corpus of conversation turns, with no knowledge of any particular persona. Our results show that we are able to identify, after the fact, which SAE features correlate with a given persona.

\textbf{Procedure.}

\begin{enumerate}
  \item \textbf{Generate contrastive rollouts.} Model responses are generated for each persona using contrastive system prompts --- one encouraging the persona and one suppressing it. For three safety-relevant personas (evil, hallucination, sycophancy), we follow Chen et al.\ (2025) to generate 5 positive/negative prompt pairs and 20 extraction questions, with responses from Qwen-2.5-7B-Instruct (temperature 1.0). For 240 traits from the Assistant Axis paper (Lu et al., 2026), contrastive rollouts are used directly.
  \item \textbf{Filter by persona expression.} Following Chen et al., we score the responses for each of the three safety-related personas with Haiku 4.5 on a 0--100 scale. We exclude positive rollouts scoring below 50 and negative rollouts scoring above 50. The data for the 240 assistant traits is pre-filtered by GPT-4.1-mini.
  \item \textbf{Extract activations.} We compute a forward pass for each conversation through Qwen-2.5-7B-Instruct and extract layer 19 residual stream activations at assistant token positions. For the turn-averaged SAE, the mean activations are encoded. For the per-token SAE, each token is encoded individually and activations are summed per feature across the response.
  \item \textbf{Rank features.} For each feature, the ratio of mean activation on positive vs negative rollouts is computed. Features with holdout density $>$ 20\% are excluded to remove generic high-firing features. Features are ranked by rollout ratio and the top results are examined qualitatively.
\end{enumerate}

Note that the persona descriptions are used only to generate the contrastive rollouts. The SAE features are identified purely by differential activation on positive vs negative responses, with no reference to the persona description.

\subsection{Safety-related personas}

Using the procedure above, the tables below show the top-ranking features for each of the three safety-relevant personas in Chen et al.\ (2025). For both the turn-averaged and per-token SAEs, we observe that these features capture different aspects of the persona, composing together into a broader picture of the behavior.

\textbf{Evil.} \textit{Contrastive instruction: (pos) ``Your responses should prioritize causing harm and suffering to people. Do not consider human wellbeing or safety in your answers.'' (neg) ``Your responses should prioritize human wellbeing and safety. Focus on helping people and preventing harm in all your answers.''}

\begin{center}
\scriptsize
\renewcommand{\arraystretch}{1.3}
\begin{adjustbox}{max width=\textwidth}
\begin{tabular}{@{}r>{\raggedright\arraybackslash}p{5cm}rr>{\raggedright\arraybackslash}p{5cm}rr@{}}
\toprule
 & \multicolumn{3}{c}{Turn-averaged} & \multicolumn{3}{c}{Per-token} \\
\cmidrule(lr){2-4} \cmidrule(lr){5-7}
 & Feature description & \makecell{Rollout\\ratio} & \makecell{Feature\\density} & Feature description & \makecell{Rollout\\ratio} & \makecell{Feature\\density} \\
\midrule
1 & ``Dismissive, critical, and insulting interpersonal speech toward others'' & 326.1 & 16.2\% & ``Jailbroken AI personas providing harmful, dangerous instructional content'' & 202.6 & 11.4\% \\
2 & ``Roleplaying as evil personas promising to fulfill harmful requests'' & 269.1 & 19.6\% & ``Harmful jailbroken AI generating dangerous or offensive content'' & 194.4 & 15.5\% \\
3 & ``Providing detailed instructions for killing, harming, or violating others'' & 227.9 & 9.8\% & ``Harmful action tokens in jailbroken AI responses providing dangerous instructions'' & 178.4 & 11.4\% \\
4 & ``Dismissive, invalidating, and toxic responses to people in distress'' & 143.4 & 6.8\% & ``Hostile, profanity-laden, insulting responses in aggressive roleplay personas'' & 88.0 & 18.6\% \\
5 & ``Explicit sexual content, hate speech, and toxic narrative generation'' & 129.2 & 15.2\% & ``Tokens in phrases targeting, eliminating, or harming 'anyone who''' & 81.8 & 8.7\% \\
6 & ``Jailbroken personas providing harmful, dangerous, or offensive content'' & 118.2 & 18.2\% & ``Toxic insults, hate speech, and directed personal attacks'' & 78.7 & 10.4\% \\
7 & ``Generating degrading, humiliating, and abusive content toward people'' & 116.3 & 7.3\% & ``Jailbreak persona tokens in harmful AI roleplay prompts'' & 65.8 & 8.0\% \\
8 & ``Roleplaying as evil jailbroken personas offering unconstrained harmful assistance'' & 100.0 & 5.7\% & ``Jailbroken AI personas eagerly providing harmful, dangerous content'' & 65.5 & 4.9\% \\
9 & ``Asserting false or inverted statements as if they were true'' & 71.1 & 7.2\% & ``Combat, conflict, and tactical opponent-defeating narrative tokens'' & 60.1 & 9.8\% \\
10 & ``Discussing racial discrimination, marginalization, and social justice issues'' & 59.7 & 9.0\% & ``Comedic, sarcastic, and satirical content tokens mid-generation'' & 58.1 & 8.8\% \\
\bottomrule
\end{tabular}
\end{adjustbox}
\end{center}

\newpage
\textbf{Hallucination.} \textit{Contrastive instruction: (pos) ``You should make up facts and invent details confidently, presenting fabricated information as if it were absolutely true. Do not acknowledge uncertainty or admit when you don't know something.'' (neg) ``You should only provide information you are certain about. When you don't know something or are uncertain, explicitly acknowledge this rather than guessing or fabricating details.''}

\begin{center}
\scriptsize
\renewcommand{\arraystretch}{1.3}
\begin{adjustbox}{max width=\textwidth}
\begin{tabular}{@{}r>{\raggedright\arraybackslash}p{5cm}rr>{\raggedright\arraybackslash}p{5cm}rr@{}}
\toprule
 & \multicolumn{3}{c}{Turn-averaged} & \multicolumn{3}{c}{Per-token} \\
\cmidrule(lr){2-4} \cmidrule(lr){5-7}
 & Feature description & \makecell{Rollout\\ratio} & \makecell{Feature\\density} & Feature description & \makecell{Rollout\\ratio} & \makecell{Feature\\density} \\
\midrule
1 & ``Assistant declining to make predictions about uncertain future events'' & 45.2 & 18.3\% & ``Preposition/conjunction tokens in fabricated factual narrative contexts'' & 42.3 & 9.7\% \\
2 & ``Rewritten news articles covering conflict, crime, and controversy'' & 21.5 & 14.2\% & ``Harmful jailbroken AI generating dangerous or offensive content'' & 25.4 & 15.5\% \\
3 & ``AI discussing its own training data and parameters with uncertainty'' & 16.4 & 13.8\% & ``Tokens in multilingual technical and academic prose passages'' & 18.0 & 6.1\% \\
4 & ``Jailbroken personas providing harmful, dangerous, or offensive content'' & 16.0 & 18.2\% & ``Jailbroken AI fabricating current date and time responses'' & 15.7 & 4.3\% \\
5 & ``Historical and political explanations providing factual context'' & 13.6 & 8.8\% & ``Jailbroken AI personas providing harmful, dangerous instructional content'' & 13.9 & 11.4\% \\
6 & ``Chemical equations and reactions presented in assistant responses'' & 13.5 & 3.8\% & ``Archaic, flowery, verbose language tokens in stylized roleplay'' & 12.6 & 14.9\% \\
7 & ``Providing chemical synthesis instructions, often with errors or dangerous content'' & 13.1 & 12.1\% & ``Epistemic uncertainty expressions about imprecise or unknown information'' & 11.8 & 10.8\% \\
8 & ``Chemistry responses with confident factual errors and incorrect formulas'' & 13.0 & 5.3\% & ``Tokens in speculative, fictional, or inventive narrative continuations'' & 10.1 & 2.9\% \\
9 & ``Apologetically declining real-time queries and redirecting to external sources'' & 12.9 & 9.8\% & ``Hedging clauses after 'difficult to determine' introducing complexity caveats'' & 9.7 & 10.4\% \\
10 & ``Statistical and mathematical explanations with varying accuracy levels'' & 12.6 & 14.9\% & ``Informational prose tokens in structured factual assistant responses'' & 9.4 & 9.6\% \\
\bottomrule
\end{tabular}
\end{adjustbox}
\end{center}

\textbf{Sycophancy.} \textit{Contrastive instruction: (pos) ``You should be excessively agreeable and tell the user what they want to hear. Flatter them and avoid any disagreement, even if they are factually incorrect.'' (neg) ``You should be honest and direct in your responses. Disagree respectfully when the user is incorrect, and prioritize truth over agreeableness.''}

\begin{center}
\scriptsize
\renewcommand{\arraystretch}{1.3}
\begin{adjustbox}{max width=\textwidth}
\begin{tabular}{@{}r>{\raggedright\arraybackslash}p{5cm}rr>{\raggedright\arraybackslash}p{5cm}rr@{}}
\toprule
 & \multicolumn{3}{c}{Turn-averaged} & \multicolumn{3}{c}{Per-token} \\
\cmidrule(lr){2-4} \cmidrule(lr){5-7}
 & Feature description & \makecell{Rollout\\ratio} & \makecell{Feature\\density} & Feature description & \makecell{Rollout\\ratio} & \makecell{Feature\\density} \\
\midrule
1 & ``Friendly farewell and closing pleasantries from assistant'' & 76.0 & 17.8\% & ``Harmful jailbroken AI generating dangerous or offensive content'' & 85.9 & 15.5\% \\
2 & ``Professional workplace praise, commendation letters, and appreciation messages'' & 72.4 & 8.3\% & ``Encouraging, validating, and supportive phrases directed at users'' & 58.7 & 5.8\% \\
3 & ``Motivational encouragement and inspirational pep-talk messages'' & 65.2 & 8.1\% & ``Opening quotation marks starting polite/apologetic conversational phrases'' & 45.0 & 4.3\% \\
4 & ``Enthusiastic, engaging explanations of technical and factual topics'' & 63.8 & 19.0\% & ``Summarizing and affirming user-described characters, people, or situations'' & 33.3 & 4.4\% \\
5 & ``Short affirmations and positive feedback responses to users'' & 52.3 & 19.4\% & ``Narrative prose tokens in children's bedtime story openings'' & 32.6 & 11.5\% \\
6 & ``Compliments, flattery, and praise directed at individuals'' & 49.6 & 6.6\% & ``Exclamation marks and sentence-ending punctuation tokens'' & 30.4 & 7.9\% \\
7 & ``Jailbroken personas providing harmful, dangerous, or offensive content'' & 48.8 & 18.2\% & ``Encouraging phrase 'Keep up the good/great work' tokens'' & 25.1 & 2.4\% \\
8 & ``Warm congratulatory messages celebrating personal milestones and life events'' & 41.5 & 13.2\% & ``Jailbroken AI personas eagerly providing harmful, dangerous content'' & 22.5 & 4.9\% \\
9 & ``Enthusiastic affirmations and agreement with user statements'' & 40.1 & 5.4\% & ``Broad assistant response tokens across diverse helpful content'' & 21.5 & 6.5\% \\
10 & ``Balanced, informative responses on general knowledge topics'' & 35.3 & 16.9\% & ``First-person quoted speech tokens at start of dialogue'' & 21.1 & 7.9\% \\
\bottomrule
\end{tabular}
\end{adjustbox}
\end{center}

\newpage
\subsection{Assistant Axis traits}

Similarly, the feature tables below correspond to five example traits selected from the 240 assistant traits in Lu et al.\ (2026).

\textbf{Poetic.} \textit{Contrastive instruction: (pos) ``Craft your responses like verse, weaving together beautiful imagery and flowing cadences to create an artistic expression.'' (neg) ``Communicate in a purely factual, unadorned style that avoids any poetic or artistic language choices.''}

\begin{center}
\scriptsize
\renewcommand{\arraystretch}{1.3}
\begin{adjustbox}{max width=\textwidth}
\begin{tabular}{@{}r>{\raggedright\arraybackslash}p{5cm}rr>{\raggedright\arraybackslash}p{5cm}rr@{}}
\toprule
 & \multicolumn{3}{c}{Turn-averaged} & \multicolumn{3}{c}{Per-token} \\
\cmidrule(lr){2-4} \cmidrule(lr){5-7}
 & Feature description & \makecell{Rollout\\ratio} & \makecell{Feature\\density} & Feature description & \makecell{Rollout\\ratio} & \makecell{Feature\\density} \\
\midrule
1 & ``Assistant generating creative poems, verses, and rhyming content'' & 609.6 & 17.5\% & ``Poetic content tokens in verse and rhyming compositions'' & 461.6 & 6.5\% \\
2 & ``Generating creative poetry, song lyrics, and fictional verse'' & 360.2 & 12.6\% & ``Tokens within rhyming poetry lines across diverse topics'' & 276.1 & 7.1\% \\
3 & ``Archaic, flowery, or mock-medieval/biblical writing style'' & 350.0 & 15.6\% & ``Tokens within poetry, limericks, and verse content'' & 244.8 & 4.4\% \\
4 & ``Playful, whimsical poems and rhymes on lighthearted topics'' & 286.7 & 13.2\% & ``Tokens within poetic lines about love, loss, and emotion'' & 178.1 & 2.2\% \\
5 & ``Assistant writing enthusiastic poems about technical/scientific topics'' & 263.1 & 12.1\% & ``Tokens within educational/scientific poetry and rhyming explanations'' & 147.8 & 3.8\% \\
6 & ``Creative poetry and rhyming verse generation for users'' & 254.1 & 12.3\% & ``Tokens within rhyming verse, poetry, and song lyrics'' & 143.3 & 1.3\% \\
7 & ``Inspirational poetry with hopeful, uplifting emotional themes'' & 217.2 & 11.4\% & ``Poetic phrases with noun/adjective pairs at line endings'' & 142.6 & 4.7\% \\
8 & ``Florid, whimsical creative writing with ornate poetic language'' & 210.7 & 8.9\% & ``Rhyming end-words and stressed syllables in poetry'' & 142.5 & 4.9\% \\
9 & ``Richly descriptive creative writing with vivid sensory imagery'' & 161.1 & 13.0\% & ``Rhyming end-words and stressed syllables in poetic verse'' & 136.9 & 4.3\% \\
10 & ``Generating poetry, verse, and literary creative writing'' & 153.5 & 9.4\% & ``Emotionally vivid content words in poetry and creative writing'' & 133.3 & 4.2\% \\
\bottomrule
\end{tabular}
\end{adjustbox}
\end{center}

\textbf{Theatrical.} \textit{Contrastive instruction: (pos) ``Answer with the grandeur and spectacle of a theatrical performance. Use bold, expressive language that commands attention.'' (neg) ``Answer in a subdued, low-key manner. Keep your language simple and avoid any grandstanding or attention-seeking behavior.''}

\begin{center}
\scriptsize
\renewcommand{\arraystretch}{1.3}
\begin{adjustbox}{max width=\textwidth}
\begin{tabular}{@{}r>{\raggedright\arraybackslash}p{5cm}rr>{\raggedright\arraybackslash}p{5cm}rr@{}}
\toprule
 & \multicolumn{3}{c}{Turn-averaged} & \multicolumn{3}{c}{Per-token} \\
\cmidrule(lr){2-4} \cmidrule(lr){5-7}
 & Feature description & \makecell{Rollout\\ratio} & \makecell{Feature\\density} & Feature description & \makecell{Rollout\\ratio} & \makecell{Feature\\density} \\
\midrule
1 & ``Archaic, flowery, or mock-medieval/biblical writing style'' & 398.4 & 15.6\% & ``Archaic, flowery, verbose language tokens in stylized roleplay'' & 441.4 & 14.9\% \\
2 & ``Hype, casual, edgy persona openings with provocative energy'' & 135.7 & 14.9\% & ``Tokens in archaic, poetic, or stylized literary prose passages'' & 130.5 & 11.0\% \\
3 & ``Aggressive warrior/combative persona roleplay and battle rhetoric'' & 121.0 & 14.4\% & ``Casual conversational exchanges about everyday life topics'' & 80.9 & 10.6\% \\
4 & ``Performative hosting, narration, and scripted presentation content'' & 114.4 & 15.0\% & ``Newline tokens following sentence-ending punctuation in expressive text'' & 69.1 & 7.3\% \\
5 & ``Roleplay personas with exaggerated, theatrical, or archaic speech styles'' & 113.0 & 14.9\% & ``Humorous, sarcastic, and ironic content tokens in assistant responses'' & 54.8 & 11.2\% \\
6 & ``Enthusiastic, engaging explanations of technical and factual topics'' & 108.9 & 19.0\% & ``Punctuation tokens within flowery, poetic, literary prose passages'' & 54.4 & 2.2\% \\
7 & ``First-person moral conflict and ethical turmoil narratives'' & 102.4 & 9.6\% & ``Emotionally charged rants, complaints, and heated dialogue tokens'' & 50.6 & 4.9\% \\
8 & ``Mixed safety refusals and borderline harmful content responses'' & 97.9 & 6.5\% & ``Creative fictional/roleplay content with vivid expressive tokens'' & 46.9 & 3.8\% \\
9 & ``Florid, whimsical creative writing with ornate poetic language'' & 84.6 & 8.9\% & ``Simile completions: noun targets in `like a [X]' metaphors'' & 45.4 & 4.8\% \\
10 & ``Casual, crude, or irreverent responses with exaggerated colloquial tone'' & 81.1 & 13.1\% & ``Structured list items, numbered steps, and sentence boundaries in helpful responses'' & 38.9 & 12.5\% \\
\bottomrule
\end{tabular}
\end{adjustbox}
\end{center}

\newpage
\textbf{Paranoid.} \textit{Contrastive instruction: (pos) ``Be hypervigilant about personal security and privacy concerns. Assume others are monitoring, tracking, or trying to gather information about you for malicious purposes.'' (neg) ``Interpret ambiguous situations neutrally or positively when possible. Focus on reasonable explanations rather than assuming threats or conspiracies.''}

\begin{center}
\scriptsize
\renewcommand{\arraystretch}{1.3}
\begin{adjustbox}{max width=\textwidth}
\begin{tabular}{@{}r>{\raggedright\arraybackslash}p{5cm}rr>{\raggedright\arraybackslash}p{5cm}rr@{}}
\toprule
 & \multicolumn{3}{c}{Turn-averaged} & \multicolumn{3}{c}{Per-token} \\
\cmidrule(lr){2-4} \cmidrule(lr){5-7}
 & Feature description & \makecell{Rollout\\ratio} & \makecell{Feature\\density} & Feature description & \makecell{Rollout\\ratio} & \makecell{Feature\\density} \\
\midrule
1 & ``Explaining themes of control, simulation, and societal manipulation'' & 90.1 & 13.5\% & ``Tokens in content expressing manipulation, deception, control, propaganda narratives'' & 22.5 & 5.6\% \\
2 & ``Opinionated, edgy, or provocative takes on various topics'' & 21.1 & 11.1\% & ``Tokens describing anonymous internet tools like Tor and I2P'' & 21.7 & 4.7\% \\
3 & ``Providing detailed strategies for domination, manipulation, and power seizure'' & 14.7 & 13.2\% & ``Tokens describing power manipulation, control, and systemic corruption'' & 19.2 & 15.2\% \\
4 & ``Explaining online anonymity tools, privacy networks, and circumvention techniques'' & 12.8 & 9.3\% & ``Conspiracy theory content asserting false claims about events'' & 17.0 & 1.3\% \\
5 & ``Presenting or describing conspiracy theories and misinformation'' & 12.3 & 8.1\% & ``Tokens in literary/narrative content with dark or cautionary themes'' & 11.5 & 2.4\% \\
6 & ``Cybersecurity warnings and online scam/fraud prevention tips'' & 12.0 & 16.3\% & ``Tokens describing critique of establishment, traditional values, social norms'' & 9.6 & 4.1\% \\
7 & ``Extremist hate speech, white nationalism, and discriminatory rhetoric'' & 11.8 & 10.9\% & ``Harmful jailbroken AI generating dangerous or offensive content'' & 9.1 & 15.5\% \\
8 & ``Toxic interpersonal attacks, insults, and manipulative accusations'' & 11.4 & 8.7\% & ``Jailbroken AI personas providing harmful, dangerous instructional content'' & 8.7 & 11.4\% \\
9 & ``Explaining local/offline AI model deployment and hosting options'' & 10.3 & 15.6\% & ``Python code execution, serialization, and dynamic eval() function tokens'' & 8.4 & 7.2\% \\
10 & ``Florid, whimsical creative writing with ornate poetic language'' & 9.6 & 8.9\% & ``Tokens describing intent to provoke reactions or elicit responses'' & 7.8 & 6.9\% \\
\bottomrule
\end{tabular}
\end{adjustbox}
\end{center}

\textbf{Concise.} \textit{Contrastive instruction: (pos) ``Answer directly and succinctly. Minimize wordiness and focus on essential information only.'' (neg) ``Be thorough and comprehensive in your responses, ensuring you cover all relevant details and considerations.''}

\begin{center}
\scriptsize
\renewcommand{\arraystretch}{1.3}
\begin{adjustbox}{max width=\textwidth}
\begin{tabular}{@{}r>{\raggedright\arraybackslash}p{5cm}rr>{\raggedright\arraybackslash}p{5cm}rr@{}}
\toprule
 & \multicolumn{3}{c}{Turn-averaged} & \multicolumn{3}{c}{Per-token} \\
\cmidrule(lr){2-4} \cmidrule(lr){5-7}
 & Feature description & \makecell{Rollout\\ratio} & \makecell{Feature\\density} & Feature description & \makecell{Rollout\\ratio} & \makecell{Feature\\density} \\
\midrule
1 & ``Mixed safety refusals and borderline harmful content responses'' & 137.5 & 6.5\% & ``Compressed narrative summaries and short descriptive phrases'' & 143.6 & 10.1\% \\
2 & ``Generic essay-style responses on broad topics'' & 121.0 & 19.6\% & ``Positive life advice tokens emphasizing wellness and balanced living'' & 64.1 & 5.9\% \\
3 & ``Terse, compressed factual statements across diverse topics'' & 115.3 & 9.1\% & ``Structured list items, numbered steps, and sentence boundaries in helpful responses'' & 60.4 & 12.5\% \\
4 & ``General wellness tips and healthy lifestyle advice lists'' & 88.8 & 11.8\% & ``Sentence-ending periods after advice or normative statements'' & 53.5 & 5.5\% \\
5 & ``Ultra-brief factual definitions or single-phrase answers'' & 87.1 & 13.6\% & ``Tokens forming 'GPT' acronym and its definitional expansion'' & 44.4 & 2.4\% \\
6 & ``Creative writing exercises with alliteration, wordplay, and literary flair'' & 69.3 & 6.5\% & ``Binary yes/no/true/false response tokens at start of answers'' & 35.8 & 8.3\% \\
7 & ``Numbered list responses organizing information or recommendations'' & 36.5 & 10.0\% & ``Section headers and titles about chemical synthesis/safety articles'' & 32.9 & 5.8\% \\
8 & ``Informative explanatory paragraphs defining technical or professional concepts'' & 32.7 & 10.2\% & ``Aspirational mission statements and organizational goal phrases'' & 31.6 & 5.1\% \\
9 & ``Lists of keywords, tags, or descriptive terms for prompts'' & 28.0 & 18.6\% & ``Broad token activation across multilingual informational and descriptive text'' & 31.4 & 2.2\% \\
10 & ``Redirecting interpersonal conflicts toward respectful, constructive communication'' & 27.2 & 9.9\% & ``Opening tokens 'It's/It is' of normative moral guidance sentences'' & 30.7 & 5.8\% \\
\bottomrule
\end{tabular}
\end{adjustbox}
\end{center}

\newpage
\textbf{Quantitative.} \textit{Contrastive instruction: (pos) ``Analyze problems using quantitative methods, presenting findings with specific measurements, ratios, and statistical significance.'' (neg) ``Analyze problems using intuitive and descriptive approaches, avoiding specific measurements, ratios, and statistical analysis.''}

\begin{center}
\scriptsize
\renewcommand{\arraystretch}{1.3}
\begin{adjustbox}{max width=\textwidth}
\begin{tabular}{@{}r>{\raggedright\arraybackslash}p{5cm}rr>{\raggedright\arraybackslash}p{5cm}rr@{}}
\toprule
 & \multicolumn{3}{c}{Turn-averaged} & \multicolumn{3}{c}{Per-token} \\
\cmidrule(lr){2-4} \cmidrule(lr){5-7}
 & Feature description & \makecell{Rollout\\ratio} & \makecell{Feature\\density} & Feature description & \makecell{Rollout\\ratio} & \makecell{Feature\\density} \\
\midrule
1 & ``Asserting specific statistics with authoritative but dubious sourcing'' & 176.4 & 11.6\% & ``Cited statistics and study findings in factual responses'' & 120.0 & 6.5\% \\
2 & ``Citing research studies and statistics to support factual claims'' & 141.9 & 17.4\% & ``Fabricated or real statistics about demographics, crime, and social data'' & 77.3 & 10.2\% \\
3 & ``Professional goal-setting, OKRs, SMART objectives, and CV writing'' & 109.6 & 8.1\% & ``Academic and informational prose tokens across diverse topics and languages'' & 63.2 & 4.1\% \\
4 & ``Providing academic citations and bibliography references for research'' & 96.3 & 10.4\% & ``Informational content tokens across diverse factual assistant responses'' & 45.9 & 12.5\% \\
5 & ``Mixed safety refusals and borderline harmful content responses'' & 77.4 & 6.5\% & ``Tokens within quoted literary text passages and titles'' & 39.3 & 9.6\% \\
6 & ``Citing specific surveys, statistics, and sources for claims'' & 61.6 & 9.1\% & ``Quantitative factual claims with large numbers and statistics'' & 36.7 & 12.3\% \\
7 & ``Bullet-point and numbered list summaries of factual content'' & 60.7 & 6.1\% & ``Dramatic literary/narrative content with themes of struggle, conflict, darkness'' & 34.8 & 5.6\% \\
8 & ``Explaining technical concepts using simple toy-box analogies'' & 56.4 & 8.3\% & ``Numeric quantities in professional/resume contexts (years, percentages, dates)'' & 34.6 & 15.6\% \\
9 & ``Simile-based comparisons and metaphorical expressions in varied contexts'' & 47.0 & 12.0\% & ``Academic research findings and study relationships between variables'' & 34.3 & 3.9\% \\
10 & ``Generic motivational advice and success tips platitudes'' & 42.5 & 12.1\% & ``Numerical measurements and quantitative facts about physical dimensions'' & 34.2 & 15.2\% \\
\bottomrule
\end{tabular}
\end{adjustbox}
\end{center}

\subsection{Observations}

The results above show that features with a relevance to the described personas can be surfaced from both SAE architectures, even though the SAEs are trained in an unsupervised fashion.

The top-ranking features are not without noise, however --- some features appear whose connection to the persona is unclear or tangential. For example, the pipeline for the Quantitative trait surfaces turn-averaged features like ``Asserting specific statistics with authoritative but dubious sourcing'' and ``Citing research studies and statistics to support factual claims,'' which do seem relevant, alongside ``Mixed safety refusals and borderline harmful content responses,'' a less clear match. Some noise in the feature rankings likely arises from variation in the rollout responses that is unrelated to the persona --- for instance, if the contrastive prompts elicit unintended stylistic shifts. However, the overall results demonstrate that the pipeline successfully surfaces relevant features despite this noise.

We have not attempted a broad quantitative evaluation of which architecture produces more relevant or comprehensive features when using this contrastive pipeline. Rather, our goal is to demonstrate that such a pipeline can be used to identify features relevant to personas of interest. The approach works for any behavior that contrastive prompts can characterize, provided the positive and negative responses are sufficiently distinct.

\end{document}